\documentclass{CUP-JNL-DCE}
\usepackage{latexsym}
\usepackage{graphicx}
\usepackage{multicol,multirow}
\usepackage{amsmath,amssymb,amsfonts}
\usepackage{mathrsfs}
\usepackage{amsthm}
\usepackage{apacite}
\usepackage{rotating}
\usepackage{appendix}
\usepackage[authoryear]{natbib}
\usepackage{ifpdf}
\usepackage[T1]{fontenc}
\usepackage{times}
\usepackage{sourcesanspro}
\usepackage{newtxmath}
\usepackage{textcomp}%
\usepackage{xcolor}%
\usepackage{hyperref}
\usepackage{subfig}
\usepackage{bm}

\articletype{SYSTEMATIC REVIEWS}

\begin{document}

\begin{Frontmatter}

\title[]
{Scalable algorithms for physics-informed neural and graph networks}

\author[1]{Khemraj Shukla} 
\author[1,2]{Mengjia Xu}
\author[3]{Nathaniel Trask}
\author*[1]{George Em Karniadakis}\email{george_karniadakis@brown.edu}

\authormark{Khemraj Shukla \textit{et al.}}

\address*[1]{\orgdiv{Division of Applied Mathematics}, \orgname{Brown University}, \orgaddress{\street{182 George St}, \state{Providence}, \postcode{RI 02912}, \country{USA}}}

\address[2]{\orgdiv{McGovern Institute for Brain Research}, \orgname{Massachusetts Institute of Technology}, \orgaddress{\street{77 Massachusetts Ave}, \state{Cambridge}, \postcode{MA 02139}, \country{USA}}}

\address[3]{\orgdiv{Center for Computing Research}, \orgname{Sandia National Laboratories}, \orgaddress{\street{1451 Innovation Pkwy SE \#600}, \state{Albuquerque}, \postcode{NM 87123}, \country{USA}}}

\received{XX-XX XX}
\revised{XX-XX-XX}
\accepted{XX-XX-XX}

\keywords{graph neural networks; PINNs; graph calculus; scalability; scientific machine learning}

\abstract{Physics-informed machine learning (PIML) has emerged as a promising new approach for simulating complex physical and biological systems that are governed by complex multiscale processes for which some data are also available. In some instances, the objective is to discover part of the hidden physics from the available data, and PIML has been shown to be particularly effective for such problems for which conventional methods may fail. Unlike commercial machine learning where training of deep neural networks requires big data, in PIML big data are not available. Instead, we can train such networks from additional information obtained by employing the physical laws and evaluating them at random points in the space-time domain. Such physics-informed machine learning integrates  multimodality and multifidelity data with mathematical models, and implements them using neural networks or graph networks. 
Here, we review some of the prevailing trends in embedding physics into machine learning, 
using physics-informed neural networks (PINNs) based primarily on feed-forward neural networks and automatic differentiation. For more complex systems or systems of systems and unstructured data, graph neural networks (GNNs) present some distinct advantages, and here we review how physics-informed learning can be accomplished with GNNs based on graph exterior calculus to construct differential operators; we refer to these architectures as physics-informed graph networks (PIGNs). We present representative examples for both forward and inverse problems and discuss what advances are needed to scale up PINNs, PIGNs and more broadly GNNs for large-scale engineering problems. \vspace*{-0.8cm}}

\begin{policy}[Impact Statement]
Many complex problems in computational engineering can be described by parametrized differential equations while boundary conditions or material properties may not be fully known, e.g., in thermal-fluid or solid mechanics systems. These ill-defined problems cannot be solved with standard numerical methods but if some sparse data are available, progress can be made by recasting them as large-scale minimization problems. Here, we review physics-informed neural networks (PINNs) and physics-informed graph networks (PIGNs) that integrate seamlessly data and mathematical physics models, even in partially understood or uncertain contexts. Using automatic differentiation in PINNs and external graph calculus in PIGNs, the physical laws are enforced by penalizing the residuals on random points in the space-time domain for PINNs and on the nodes of a graph in PIGNs. New multi-GPU algorithms are needed to scale up PINNs and PIGNs to realistic applications.
\end{policy}
\end{Frontmatter}

\section{Introduction}
\label{sec:Introduction} 
Physics-informed machine learning involves the use of neural networks, graph networks or Gaussian process regression to simulate physical and biomedical systems, using a combination of mathematical models and multi-modality data
\citep{karniadakis2021physics,raissi2019physics,gao2022physics,raissi2018numerical}. 

\begin{figure}[b!]
\centering
\includegraphics[scale=0.35]{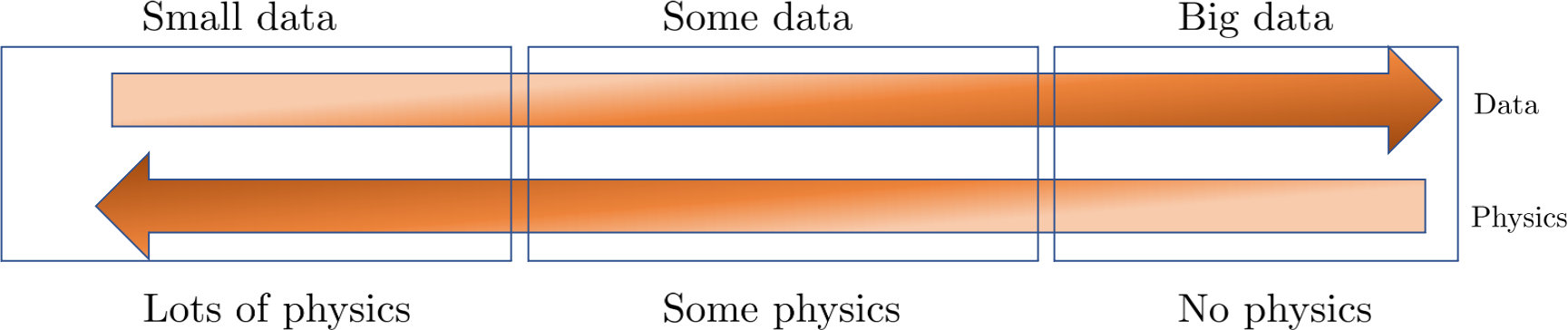}
\caption{\small {\textbf{Physics versus Data.} Simulating and forecasting the response of real-world problems requires both data and physical models. In many applications throughout physics, engineering and biomedicine we have some data and we can describe some but not all physical process. Physics-informed machine learning enables seamless integration of data and models. On the left, we have the classical paradigm of well posed problems. On the right, we depict black-box type system identification. Most real-life applications fall in the middle of this diagram.}}
\label{fig:Data_Physics}
\end{figure}

Physics-informed neural networks (PINNs)~\citep{raissi2019physics} can solve a partial differential equation (PDE) by directly incorporating the PDE into the loss function of the neural network (NN) and employing automatic differentiation to represent all the differential operators. Hence, PINNs do not require any mesh generation, thus avoiding a tremendous cost that has hindered progress in computational engineering, especially for moving and deformable domains. They can easily be applied to any known PDE, but also to different types of differential equations, including fractional PDEs~\citep{pang2019fpinns}, integro-differential equations~\citep{lu2021deepxde}, or stochastic PDEs~\citep{zhang2019quantifying,zhang2020learning}. PINNs are very simple to implement, and even the most elaborate codes contain less than 1,000 lines. Unlike traditional numerical approaches, the same exact code is able to treat both forward and inverse problems. Hence, different simulation scenarios, as shown in Fig. \ref{fig:Data_Physics}, can be executed without any extra effort, other than preparing the input data.
These advantages have been demonstrated across different fields, e.g., in fluid mechanics~\citep{raissi2020hidden,mao2020physics}, optics~\citep{chen2020physics,lu2021physics}, systems biology~\citep{yazdani2020systems}, geophysics \citep{waheed2020eikonal}, non-destructive evaluation of materials \citep{shukla2020physics,  shukla2021physics}, and biomedicine~\citep{sahli2020physics,yin_thrombus2020,drugs_PINNs}.
%

\begin{figure}
\centering
\includegraphics[scale=0.36]{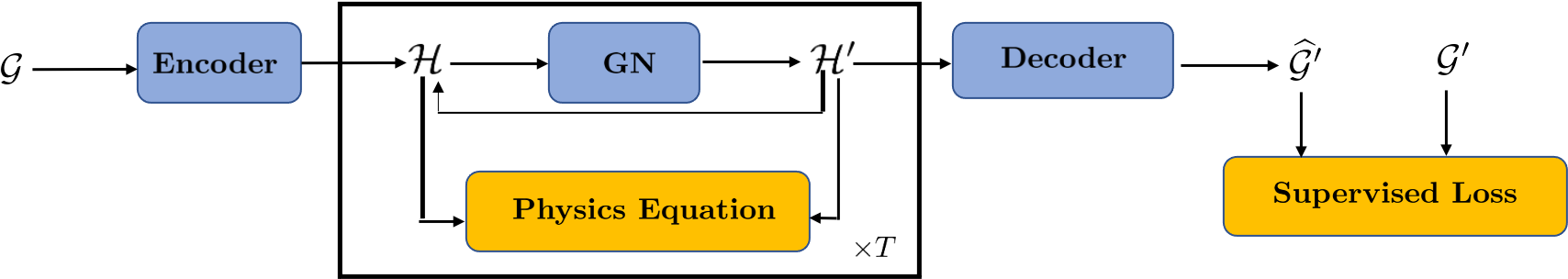}
\caption{\small {\textbf{Recurrent architecture to incorporate physics in graph networks.} The blue blocks contain learnable parameters while the orange blocks
are objective functions. The middle core block can be repeated as many times
as the required time steps (T). Schematic adopted from \citep{seo2019differentiable}.}} 
\label{fig:DPGN}
\end{figure}

Graph neural networks (GNNs) and specifically physics-informed graph networks (PIGNs) have emerged as an alternative method, as they can be particularly effective for composite physical systems where a state evolves as a function of its neighboring states, forming dynamic relational graphs instead of grids. In contrast to PINNs, these models apply message-passing between a small number of moving and interacting objects, which deviate from PDEs that are strictly differential functions. Another application of GNNs and PIGNs is the discovery of 
PDEs from sparse data, see e.g., \citep{iakovlev2020learning}. The authors combined message passing neural networks with the method of lines and neural ODE \citep{Chen2018} and obtained good results for advection-diffusion, heat equation and Burgers equation in one dimension. Similar work was done in \citep{kumar2021grade}, where the authors designed a PIGN called GrADE 
for learning the system dynamics from data, e.g., the one-dimensional and two-dimensional Burgers equations. GrADE combines a Graph Neural Network (GNN) that can handle unstructured data with neural ODE to represent the temporal domain.
In addition, the authors used the graph attention (GAT) for embedding of nodes during aggregation.
A somewhat different approach was proposed in \citep{gao2022physics}, where the authors
introduced the graph convolution operation into physics-informed learning and
combined it with finite elements in order to handle unstructured meshes for fluid mechanics and solid mechanics applications. 
Graph neural networks have also been proposed for solving ordinary differential equations (GNODE) \citep{Poli_2019} in modeling continuous-time signals on graphs. 
This work was extended in \citep{seo2019differentiable} for PDEs and for applications to climate, where a recurrent architecture was proposed to incorporate physics on graph networks, see Fig. \ref{fig:DPGN}. We note that in this architecture  the physics is not directly applied to the input observations but rather to the latent representations.

One of the major limitations of the original PINN algorithm is the large computational cost associated with the training of the neural networks, especially for forward multi-scale problems. To reduce the computational cost,  \cite{jagtap2020conservative} introduced a domain decomposition-based PINN for conservation laws, namely conservative PINN (cPINN), where continuity of the states as well as their fluxes across subdomain interfaces is enforced to connect the subdomains together. In subsequent work, \cite{jagtap2020extended} applied domain decomposition to general PDEs using the so-called extended PINN (XPINN). Unlike cPINN, which offers space decomposition, XPINN offers both space-time domain decomposition for any irregular, non-convex geometry thereby reducing the computational cost effectively. By exploiting the decomposition process of the cPINN/XPINN methods and its implicit mapping onto modern heterogeneous computing platforms, the training time of the network can be reduced to a great extent. Another domain decomposition method applied to the variational formulation of PINNs was proposed in \citep{hpVPINN}. Referred to as the \textit{hp-VPINN} method, it mimics the well-known spectral element method \citep{karniadakis2005spectral}, which exhibits dual {\em h-p} convergence. An effective parallel implementation of PINNs was reported in \citep{shukla2021parallel}. Similarly, drawing on existing parallel frameworks for standard graph neural networks (GNNs), one can scale up PIGNs for simulating complex systems or systems. 

This chapter is organized as follows. In the next section we review the PINN formulation and the scalable extensions, and provide prototypical examples. In the subsequent section, we review popular variants of GNNs (spectral, spatial and graph-attention) before discussing their relationship to traditional PDE discretizations and introducing PIGNs, which employ a graph exterior calculus; we also discuss scalability of GNNs. 

%

%
\section{Physics-Informed Neural Networks (PINNs)}
\label{sec:PINNs}

We consider the following PDE for the solution $u(\mathbf{x},t)$ parametrized by the parameters $\bm{\lambda}$ defined on a domain $\Omega$:
\begin{equation} \label{eq:pde}
    f\left(\mathbf{x} ; \frac{\partial u}{\partial x_{1}}, \ldots, \frac{\partial u}{\partial x_{d}} ; \frac{\partial^{2} u}{\partial x_{1} \partial x_{1}}, \ldots, \frac{\partial^{2} u}{\partial x_{1} \partial x_{d}} ; \ldots ; \boldsymbol{\lambda}\right)=0, \quad \mathbf{x}=(x_1, \cdots,x_d) \in \Omega,
\end{equation}
with the boundary conditions
$$\mathcal{B}(u, \mathbf{x})=0 \quad \text { on } \quad \partial \Omega.$$
In PINNs, the initial conditions are treated similarly as the boundary conditions.

\begin{figure}
\centering
\includegraphics[scale=0.34]{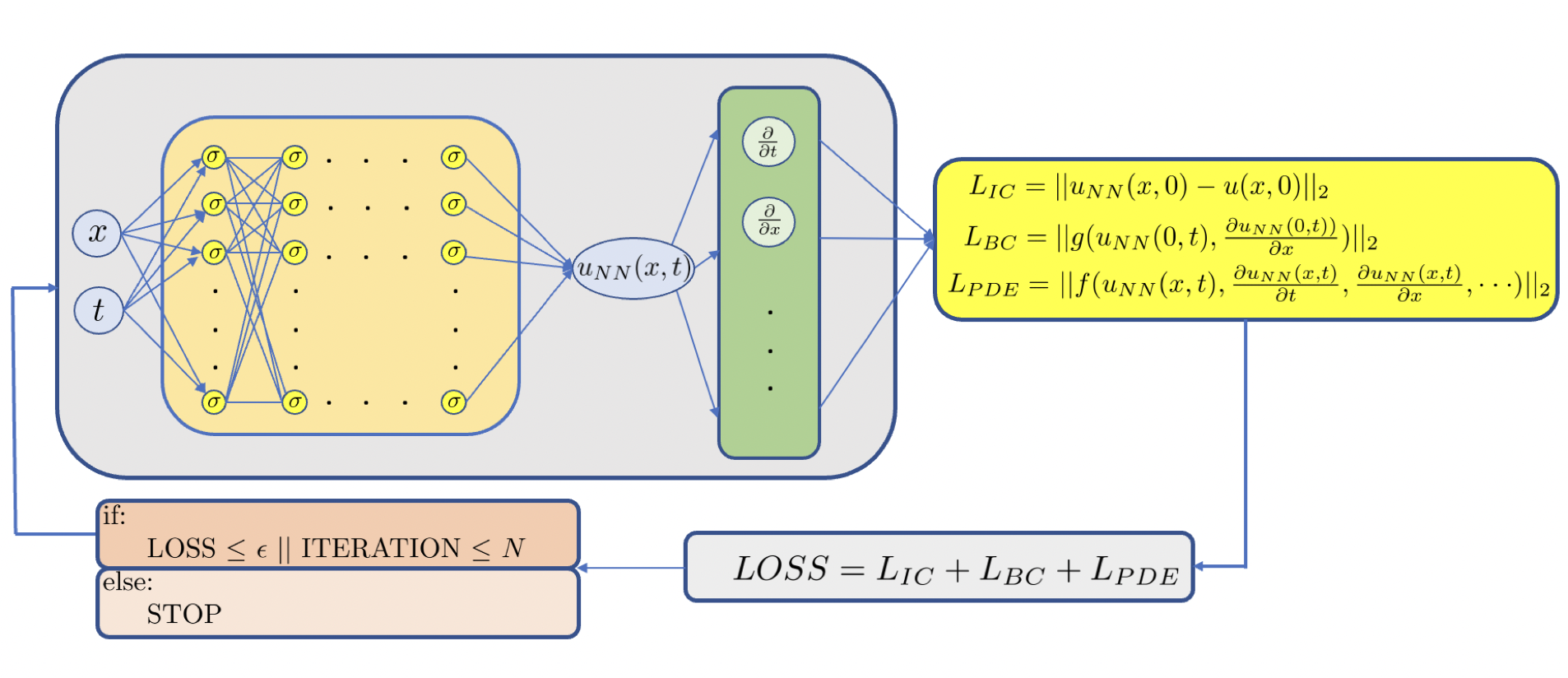}
\caption{\small {\textbf{Schematic of PINN.} The left part represents the data NN whereas the right part represents the physics-informed NN. All differential operators are obtained via automatic differentiation, hence no mesh generation is required to solve the PDE.}} 
\label{fig:PINN}
\end{figure}

A schematic of a PINN for Eq. \ref{eq:pde} is shown in Fig. \ref{fig:PINN}.
Specifically, in order 
to solve the parametrized PDE via PINNs, we construct a deep NN $\hat{u}(\mathbf{x}; \boldsymbol{\theta})$ with the trainable parameters $\boldsymbol{\theta}$.
We incorporate the constraints implied by the PDE as well as the the boundary and initial conditions in the loss function. To evaluate the residual of the PDE and the boundary/initial  conditions, we introduce a set of random points inside the domain ($\mathcal{T}_f$) and another random set of points on the boundary ($\mathcal{T}_b$). The loss function is then defined as~\citep{raissi2019physics,lu2021deepxde}
$$\mathcal{L}(\boldsymbol{\theta} ; \mathcal{T})=w_{f} \mathcal{L}_{f}\left(\boldsymbol{\theta} ; \mathcal{T}_{f}\right)+w_{b} \mathcal{L}_{b}\left(\boldsymbol{\theta} ; \mathcal{T}_{b}\right),$$
where
\begin{equation} \label{eq:loss_f}
    \mathcal{L}_{f}\left(\boldsymbol{\theta} ; \mathcal{T}_{f}\right) = \frac{1}{\left|\mathcal{T}_{f}\right|} \sum_{\mathbf{x} \in \mathcal{T}_{f}} \left|f\left(\mathbf{x};\frac{\partial \hat{u}}{\partial x_{1}}, \ldots, \frac{\partial \hat{u}}{\partial x_{d}} ; \frac{\partial^{2} \hat{u}}{\partial x_{1} \partial x_{1}}, \ldots, \frac{\partial^{2} \hat{u}}{\partial x_{1} \partial x_{d}} ; \ldots ; \boldsymbol{\lambda}\right)\right|^{2},
\end{equation}
\begin{equation} \label{eq:loss_bc}
    \mathcal{L}_{b}\left(\boldsymbol{\theta} ; \mathcal{T}_{b}\right) = \frac{1}{\left|\mathcal{T}_{b}\right|} \sum_{\mathbf{x} \in \mathcal{T}_{b}} \left|\mathcal{B}(\hat{u}, \mathbf{x})\right|^{2}.
\end{equation}
Here, $w_f$ and $w_b$ are the weights, and their choice is very important, especially for multi-scale problems; below we show an effective algorithm on how to learn these weights, which can be a function of space-time, directly from the data.

One main advantage of PINNs is that the same formulation can be used  not only for forward problems but also for inverse PDE-based problems. If
the parameter $\boldsymbol{\lambda}$ in Eq.~\eqref{eq:pde} is unknown, and instead we have some extra measurements of $u$  on the set of points $\mathcal{T}_{i}$, then we add an additional data loss~\citep{raissi2019physics,lu2021deepxde} as
$$\mathcal{L}_{i}\left(\boldsymbol{\theta}, \boldsymbol{\lambda} ; \mathcal{T}_{i}\right)=\frac{1}{\left|\mathcal{T}_{i}\right|} \sum_{\mathbf{x} \in \mathcal{T}_{i}} |\hat{u}(\mathbf{x}) - u(\mathbf{x})|^{2}$$
to learn the unknown parameters simultaneously with the solution $u$. We can recast the loss function as
$$\mathcal{L}(\boldsymbol{\theta}, \boldsymbol{\lambda} ; \mathcal{T})=w_{f} \mathcal{L}_{f}\left(\boldsymbol{\theta}, \boldsymbol{\lambda} ; \mathcal{T}_{f}\right)+w_{b} \mathcal{L}_{b}\left(\boldsymbol{\theta}, \boldsymbol{\lambda} ; \mathcal{T}_{b}\right)+w_{i} \mathcal{L}_{i}\left(\boldsymbol{\theta}, \boldsymbol{\lambda} ; \mathcal{T}_{i}\right).$$

\subsection{Self-adaptive loss weights}
 
 In some cases, it is possible to enforce the boundary conditions automatically by modifying the network architecture~\citep{lagaris1998artificial,pang2019fpinns,lagari2020systematic,lu2021physics}. A more general approach was proposed by \citep{mcclenny2020self},
 which we explain next and give a simple example of a boundary layer problem.
Let us consider a boundary layer problem defined by an ODE as follows
  \begin{align} \label{odebl}
  \nu \frac{\mathrm d^2 u}{\mathrm d x^2} - u = \text{e}^x,
\end{align}
where $x \in [-1, 1]$, $u[-1]=1$, $u[1] = 0$, and viscosity $\nu=10^{-3}$.

\begin{figure}[htbp] \label{SAPINN}
    \centering
    \includegraphics[width=0.45\textwidth]{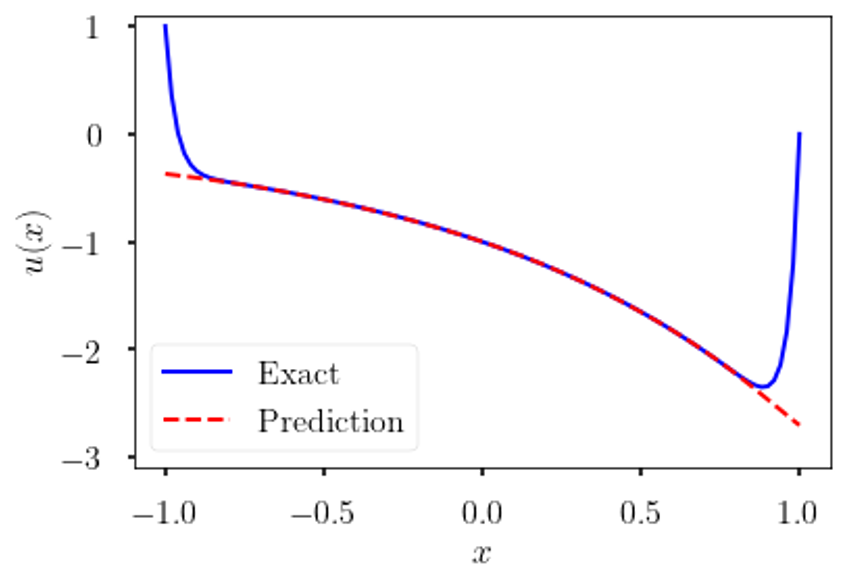}
    \includegraphics[width=0.45\textwidth]{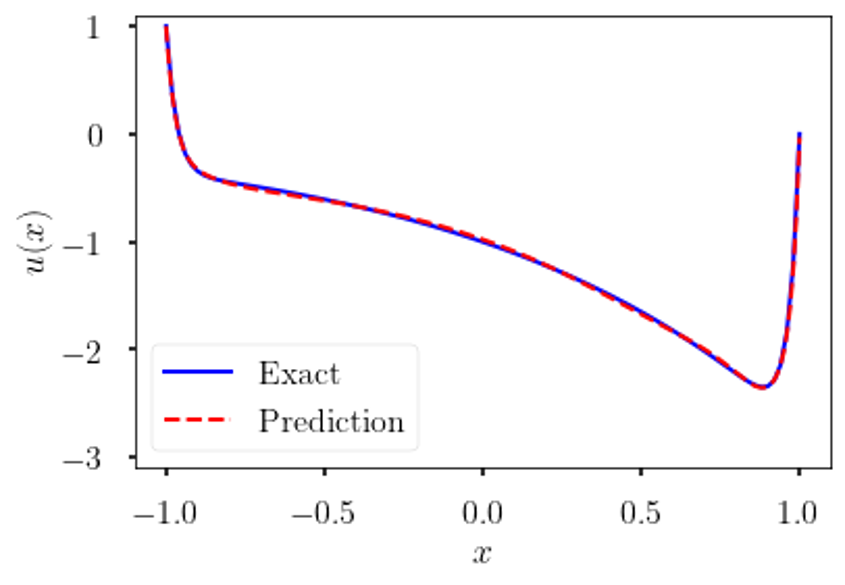}
    \caption{\small {\textbf{Self-adaptive weights.} Solution of \eqref{odebl} using vanilla PINN (left) and self-adaptive PINNs(right). The self-adaptive PINN can capture the boundary layers whereas the vanilla PINN fails.}} 
  \end{figure}
To compute the solution of \eqref{odebl}, the loss function for self-adaptive PINN is defined as
\begin{align}
\mathcal{L}\left(w, \lambda_{r}, \lambda_{b}, \lambda_{0}\right)=\mathcal{L}_{r}\left(w, \lambda_{r}\right)+\mathcal{L}_{b}\left(w, \lambda_{b}\right)+\mathcal{L}_{0}\left(w, \lambda_{0}\right),
\end{align}
where $\lambda_{r}=\left(\lambda_{r}^{1}, \ldots, \lambda_{r}^{N_{r}}\right), \lambda_{b}=\left(\lambda_{b}^{1}, \ldots, \lambda_{b}^{N_{b}}\right)$, and $\lambda_{n}=\left(\lambda_{0}^{1}, \ldots, \lambda_{0}^{N_{0}}\right)$ are trainable self-adaptation weights for the initial, boundary, and collocation points, respectively, and
\begin{align} \label{loss-sa}
\begin{aligned}
&\mathcal{L}_{r}\left(w, \lambda_{r}\right)=\frac{1}{N_{r}} \sum_{i=1}^{N_{r}}\left[\lambda_{r}^{i} r\left(x_{r^{\prime}}^{i} t_{r}^{1}: w\right)\right]^{2} \\
&\mathcal{L}_{b}\left(w, \lambda_{b}\right)-\frac{1}{N_{b}} \sum_{i=1}^{N_{b}}\left[\lambda_{b}^{i}\left(u\left(x_{b}^{i}, t_{b,}^{i} ; w\right)-g_{b}^{\prime}\right)\right]^{2} \\
&\mathcal{L}_{0}\left(w, \lambda_{0}\right)=\frac{1}{N_{0}} \sum_{i=1}^{N_{0}}\left[\lambda_{0}^{i}\left(u\left(x_{0}^{i}, 0 ; w\right)-h_{0}^{i}\right)\right]^{2}.
\end{aligned}
\end{align}

The key feature of self-adaptive PINNs is that the loss $\mathcal{L}\left(w,\lambda_r, \lambda_{b}, \lambda_{0}\right)$ is minimized with respect to the network weights $w$, as usual, but is maximized with respect to the self-adaptation weights $\lambda_{r}, \lambda_{b}, \lambda_{0}$; in other words, in training we seek a saddle point
\begin{align}\label{optim-sa}
\min _{w} \max _{r, \lambda_{b}, \lambda_{0}} \mathcal{L}\left(w, \lambda_{r}, \lambda_{b}, \lambda_{0}\right) .
\end{align}
This can be accomplished by a gradient descent-ascent procedure, with updates given by
\begin{align} \label{gd-sa}
\begin{aligned}
w^{k-1} &=w^{k}-\eta_{k} \nabla_{w} \mathcal{L}\left(w^{k}, \lambda_{r}^{k}, \lambda_{i}^{k}, \lambda_{0}^{k}\right), \\
\lambda_{r}^{k+1} &=\lambda_{r}^{k}+\eta_{k} \nabla_{\lambda \cdot} \mathcal{L}\left(w^{k}, \lambda_{r}^{k}, \lambda_{b}^{k}, \lambda_{0}^{k}\right), \\
\lambda_{b}^{k+1} &=\lambda_{b}^{k}+\eta_{k} \nabla_{\lambda_{0}} \mathcal{L}\left(w^{k}, \lambda_{r}^{k}, \lambda_{b}^{k}, \lambda_{0}^{k}\right), \\
\lambda_{0}^{k+1} &=\lambda_{0}^{k}+\eta_{k} \nabla_{\lambda_{0}} \mathcal{L}\left(w^{k}, \lambda_{r}^{k}, \lambda_{b}^{k}, \lambda_{0}^{k}\right).
\end{aligned}
\end{align}
In the case of vanilla PINN, the $\lambda$'s are positive constant scalars, which weight the solution equally and do not adapt to the regularity of the solution. However, in self-adaptive PINN, initial, boundary or collocation points in stiff regions of the solution 
automatically emphasize more these terms in the loss function, hence forcing the approximation to improve on those regions.

\subsection{Gradient-enhanced training of PINNs}

In the standard PINN algorithm we minimize the residual $f$ of the PDE in the $L_2$-norm 
but it may be beneficial to minimize it in the Sobolev norm, and this makes sense since  the derivatives of $f$ are also zero. This was proposed in \citep{yu2022gradient} in the 
gradient-enhanced PINNs (gPINNs), i.e.,
$$\nabla f(\mathbf{x}) = \left(\frac{\partial f}{\partial x_1}, \frac{\partial f}{\partial x_2}, \cdots, \frac{\partial f}{\partial x_d}\right) = \mathbf{0}, \quad \mathbf{x} \in \Omega.$$
Hence, the loss function of gPINNs is:
$$\mathcal{L} = w_{f} \mathcal{L}_{f} +w_{b} \mathcal{L}_{b} + w_i\mathcal{L}_i + \sum_{i=1}^d w_{g_i} \mathcal{L}_{g_i}\left(\boldsymbol{\theta} ; \mathcal{T}_{g_i}\right),$$
where
\begin{equation} \label{eq:loss_g}
    \mathcal{L}_{g_i}\left(\boldsymbol{\theta} ; \mathcal{T}_{g_i}\right) = \frac{1}{\left|\mathcal{T}_{g_i}\right|} \sum_{\mathbf{x} \in \mathcal{T}_{g_i}} \left| \frac{\partial f}{\partial x_{i}} \right|^{2}.
\end{equation}
Here, $\mathcal{T}_{g_i}$ is the set of residual points for the derivative $\frac{\partial f}{\partial x_{i}}$, and in general,  $\mathcal{T}_f$ and $\mathcal{T}_{g_i}$ ($i=1, \cdots, d$) can be different.

For example, for the Poisson's equation $\Delta u = f$ in 1D, the additional loss term is
$$\mathcal{L}_{g} = w_{g} \frac{1}{\left|\mathcal{T}_{g}\right|} \sum_{\mathbf{x} \in \mathcal{T}_{g}} \left|\frac{d^3 u}{d x^3} - \frac{d f}{d x} \right|^{2}.$$

Next, we present a comparison between PINNs and gPINNs based on the work of \citep{yu2022gradient} for the following diffusion equation
\begin{equation*}
    \frac{\partial u}{\partial t} = D \frac{\partial^2 u}{\partial x^2} + R(x,t), \quad x \in [-\pi, \pi],~ t \in [0, 1],
\end{equation*}
where $D = 1$.
Here, $R$ is given by: 
$$
R(x,t) = e^{-t} \left[ \frac{3}{2}\sin(2x) + \frac{8}{3}\sin(3x) + \frac{15}{4}\sin(4x) + \frac{63}{8}\sin(8x) \right].
$$

The initial and boundary conditions are:
\begin{gather*}
    u(x, 0) = \sum^{4}_{i=1} \frac{\sin(ix)}{i} + \frac{\sin(8x)}{8}, \\
    u(-\pi, t) = u(\pi, t) = 0,
\end{gather*}
corresponding to the analytical solution for $u$:
\begin{equation} \label{eq:dr_exact}
    u(x, t) = e^{-t} \left[ \sum^4_{i=1} \frac{\sin(ix)}{i} + \frac{\sin(8x)}{8} \right].
\end{equation}
We have two loss terms of the gradient, and the total loss function is
$$\mathcal{L} = \mathcal{L}_f + w \mathcal{L}_{g_x}+w \mathcal{L}_{g_t}.$$

\begin{figure}[htbp]
    \centering
    \includegraphics[width=.9\textwidth]{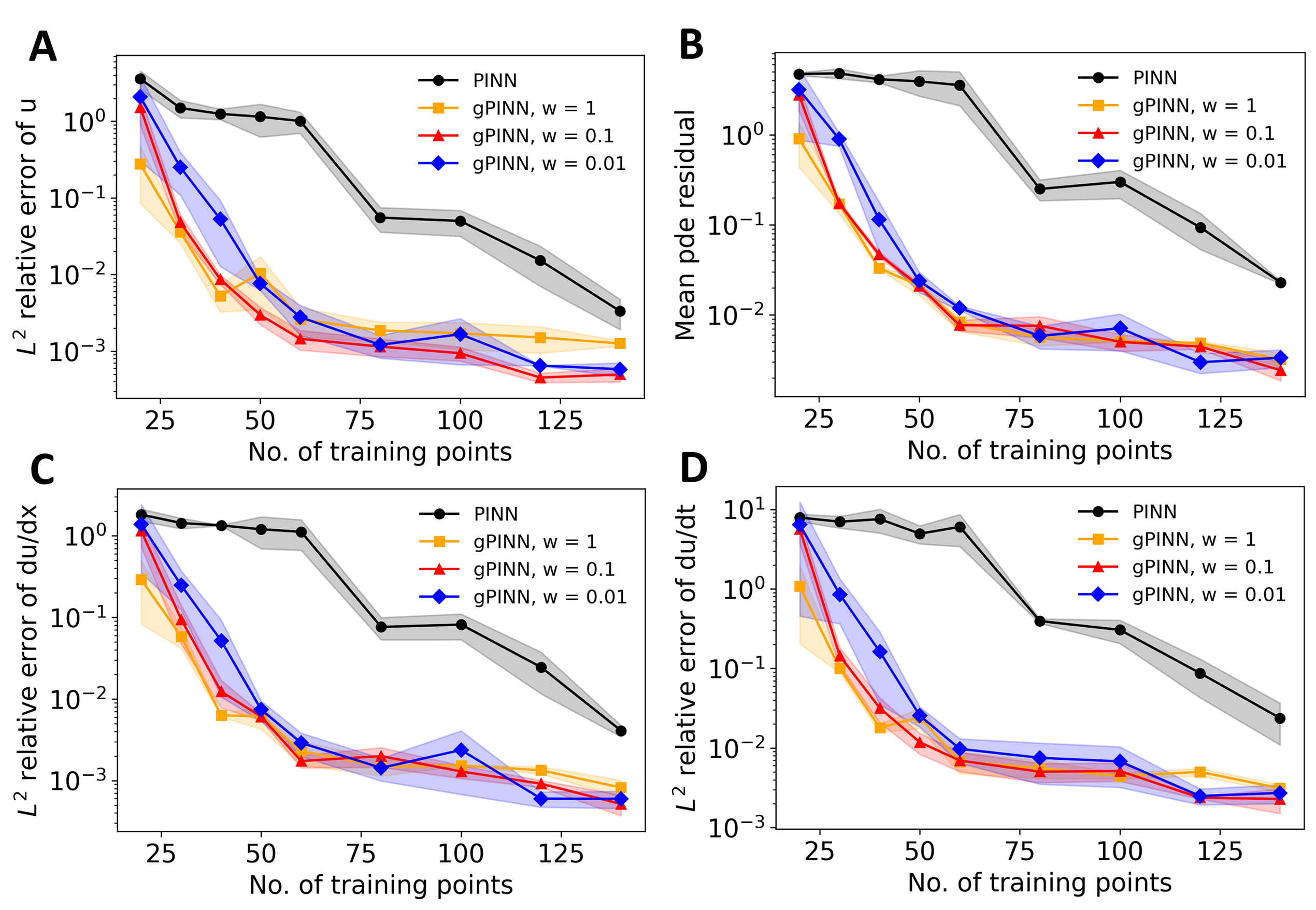}
    \caption{\small {\textbf{Comparison between PINN and gPINN.} (\textbf{A}) $L^2$ relative error of $u$ for PINN and gPINN with $w=$1, 0.1, and 0.01. (\textbf{B}) Mean absolute value of the PDE residual. (\textbf{C}) $L^2$ relative error of $\frac{du}{dx}$. (\textbf{D}) $L^2$ relative error of $\frac{du}{dt}$.
    (Figure is from \citep{yu2022gradient}).}}  
    \label{fig:3.2.2.1}
\end{figure}

We see in Fig.~\ref{fig:3.2.2.1} that gPINN results with the values of $w=0.01$, 0.1, and 1 all outperform PINN by up to two orders of magnitude.
However, we note that while gPINN reaches 1\% $L^2$ relative error of $u$ by using only 40 training points, and PINN requires more than 100 points to reach the same accuracy, gPINN is 
almost twice as expensive as PINN due to extra expensive automatic differentiation.

\subsection{Scalable PINNs}
As discussed above, PINNs can efficiently tackle both forward problems, where the solution of governing physical law is inferred, as well as ill-posed inverse problems, where the unknown physics and/or free parameters in the governing equations are to inferred from the available multi-modal measurements.
However, one of the major concerns with PINNs is the large computational cost associated with the training of the neural networks, especially for forward multi-scale problems. To improve the scalebility  and reduce the computational cost, Jagtap \etal \citep{jagtap2020conservative} introduced a domain decomposition-based (in-space) PINN for conservation laws, namely conservative PINN (cPINN), where continuity of the state variables as well as their fluxes across subdomain interfaces is used to obtain the global solution from the local solutions in the subdomains. In subsequent work, Jagtap \& Karniadakis \citep{jagtap2020extended} applied domain decomposition to general PDEs using the so-called extended PINN (XPINN). Unlike cPINN, which offers space decomposition, XPINN offers both space-time domain decomposition for any irregular, non-convex geometry thereby reducing the computational cost effectively. By exploiting the decomposition process of the cPINN/XPINN methods and its implicit mapping on the modern heterogeneous computing platforms, the training time of the network can be reduced to a great extent. 

\begin{figure}
\centering
\subfloat[] {
\includegraphics[ scale=0.25]{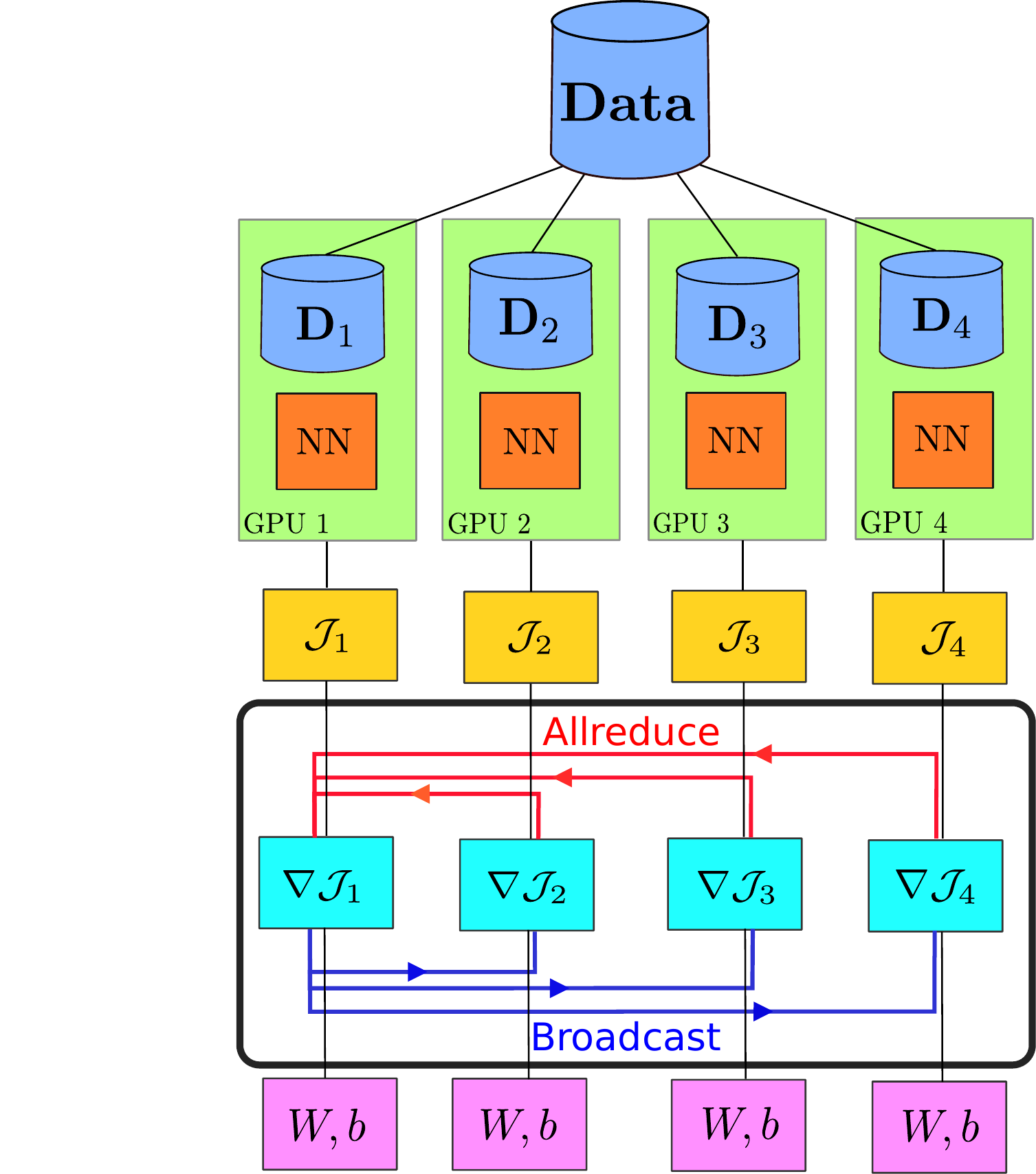}
\label{dp}
}  
\subfloat[]{
\includegraphics[ scale=0.55]{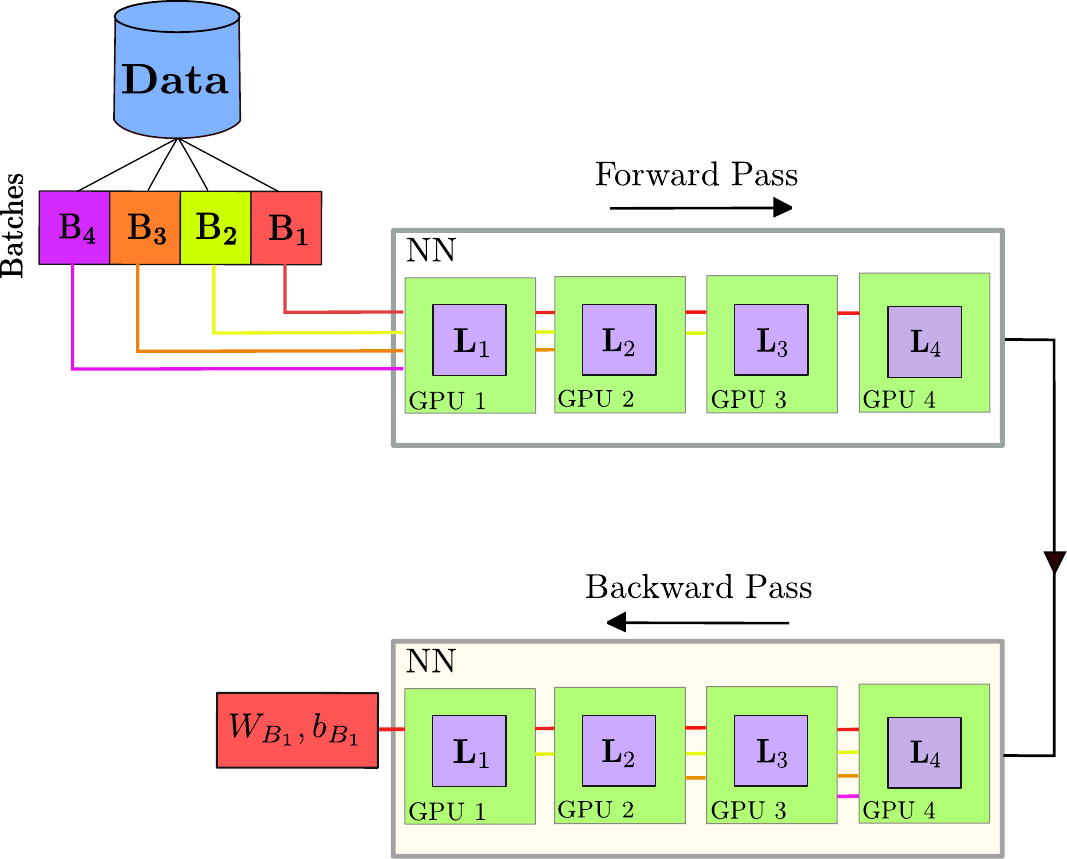}
\label{mp}
}
\caption{\small {\textbf{Schematic of the implementation of data and model parallel algorithms.} (a) Method for the data-parallel approach, where the same neural network model, represented by NN, is loaded by each processor but works on a different chunk of input data. Synchronization of training (gradient of loss) is performed after the computation of loss on each processor via "allreduce" and "broadcast" operations represented by horizontal red and blue lines. (b) represents the model parallel approach, where each layer of the model (represented by $\text{L}_1 \ldots \text{L}_4)$ is loaded on a processor and each processor works on a batch of data $(\text{B}_1 \ldots \text{B}_4)$. Forward and backward passes are performed by using a pipeline approach. Adopted from \citep{shukla2021parallel}.}\}}
\label{data_model}
\end{figure}

There are currently two existing approaches for distributed training of neural networks, namely, the data-parallel approach \citep{sergeev2018horovod} and the model parallel approach, which are agnostic to physics-based priors. The data-parallel approach is based on the single instruction and multiple data (SIMD) parallel programming model, which results in a simple performance model driven by weak scaling. A block diagram showing the basic building blocks of the data-parallel approach is shown in Fig. \ref{dp} for a four processors or co-processors system. The programming model used for the data-parallel approach falls into the regime of MPI+X system, where X is chosen as CPU(s) or GPU(s), depending on the size of the input data. In the data-parallel approach, the data is uniformly split into a number of chunks $(\text{D}_1,\ldots, \text{D}_4~\text{in Fig. \ref{dp}})$, equal to the number of processors. The neural network (NN) model is initialized with the same parameters on all the processes as shown in Fig. \ref{dp}. These neural networks are working on different chunks of the data, and therefore, work on different loss functions as shown by $\mathcal{J}_1, \ldots, \mathcal{J}_4$ in Fig. \ref{dp}.

 NVIDIA also introduced the parallel code MODULUS \citep{hennigh2020nvidia}, which implements the standard PINN-based multi-physics simulation framework. The underlying idea of MODULUS is to solve the differential equation by modeling the mass balance condition as a hard constraint as well as a global constraint. MODULUS provides the functionality for multi-GPU and multi-Node implementation based on the data-parallel approach (Fig. \ref{dp}).

To ensure a consistent neural network model (defined with same weights and biases) across all the processes and during each epoch or iteration of training, a distributed optimizer is used, which averages out the gradient of loss values $(\nabla \mathcal{J}_1,\ldots,\nabla \mathcal{J}_4~\text{in Fig. \ref{dp} })$ on  the root processor through an "allreduce" operation. Subsequently, the averaged gradient of the loss function is sent to all the processors in the communicator world through a broadcast operation (collective communication). Then, the parameters of the model are updated through an optimizer. An additional component, which arises in the data-parallel approach, is the increase in global batch size with the number of processes. Goyal \etal \citep{goyal2017accurate} addressed this issue by multiplying the learning rate with the number of processes in the communicator. We note that in the data-parallel approach the model size (size of neural network parameters) remains uniform on each processor or GPU and that imposes a problem for large models to be trained on GPUs as they have fixed memory. 

To circumvent this issue, another distributed algorithm approach namely, the model parallel approach is proposed. 
A block diagram of the algorithm in the model parallel approach is shown in Fig. \ref{mp}, which can be interpreted as a classic example of pipeline algorithm \citep{DeepSpeed}. In the model parallel approach, data is first divided into small batches $(\textbf{B}_1,\ldots,\textbf{B}_4~\text{in Fig. \ref{mp}})$ and each hidden layer $(\textbf{L}_1,\ldots,\textbf{L}_4~\text{in Fig. \ref{mp}})$ is distributed to a processor (or GPU). During the forward pass, once $\textbf{B}_1$ is processed by $\textbf{L}_1$, the output of $\textbf{L}_1$ is passed as the input to $\textbf{L}_2$ and $\textbf{L}_1$ will start working on $\textbf{B}_2$ and so on. Once $\textbf{L}_4$ finishes working, $\textbf{B}_1$ the backward pass (optimization process) will kickoff, thus, completing epochs of training in a pipeline mode. We note that the implementation of both algorithms is problem agnostic and does not incorporate any prior information on solutions to be approximated, which makes the performance of these algorithms to be dependent on the data size and model parameters. In the literature, the implementation of data and model parallel approaches is primarily carried out for problems pertaining to the classification and natural language processing \citep{goyal2017accurate, rasley2020deepspeed}, which are based on large amounts of training data. Therefore, the efficiency of data and the model parallel approach for scientific machine learning is not explored, which is primarily dominated by the high-dimensional and sparse data set. Apart from these two classical approaches, recently Xu \etal \citep{xu2020distributed} deployed the topological concurrency on data structures of neural network parameters. In brief, this implementation could be comprehended as task-based parallelism and it is rooted in the idea of model-based parallelism. Additionally, it also provides interactivity with other discretization-based solvers such as Fenics \citep{alnaes2015fenics}.

 The main advantage of cPINN and XPINN over data and model parallel approaches (including the work of Xu \etal \citep{xu2020distributed}) is their adaptivity for all hyperparameters of the neural network in each subdomain. In the vanilla data-parallel approach, the training across processors is synchronized by averaging the gradient of loss function across and subsequently broadcasting, enforcing the same network architectures on each processor. In PIML, the solution of the underlying physical laws is inferred based on a small amount of data, where the chosen neural network architecture depends on the complexity and regularity of the solutions to be recovered. To address this issue, the synchronization of the training process across all the processors has to be achieved by the physics of the problem, and therefore cPINNs and XPINNs come to the rescue. The convergence of the solution based on a single domain is constrained by its global approximation, which is relatively slow. On the other hand, computation of the solution based on local approximation by using the local neural networks can result in fast convergence. With regards to computational aspects, the domain decomposition-based approach requires a point-to-point communication protocol with a very small size of the buffer. However, for the data-parallel approach, it requires one \textit{allreduce} operation and one \textit{broadcast} operation with each having the buffer size equal to the number of parameters of the neural network, which makes communication across the processor slower. Moreover, for problems involving heterogeneous physics the data/model parallel approaches are not adequate, whereas cPINNs/XPINNs can easily handle such multi-physics problems.  Therefore, domain decomposition approaches paired with physics-based synchronization help more than the vanilla data-parallel approaches.
 
\begin{figure}
\centering
 \includegraphics[ scale=0.6]{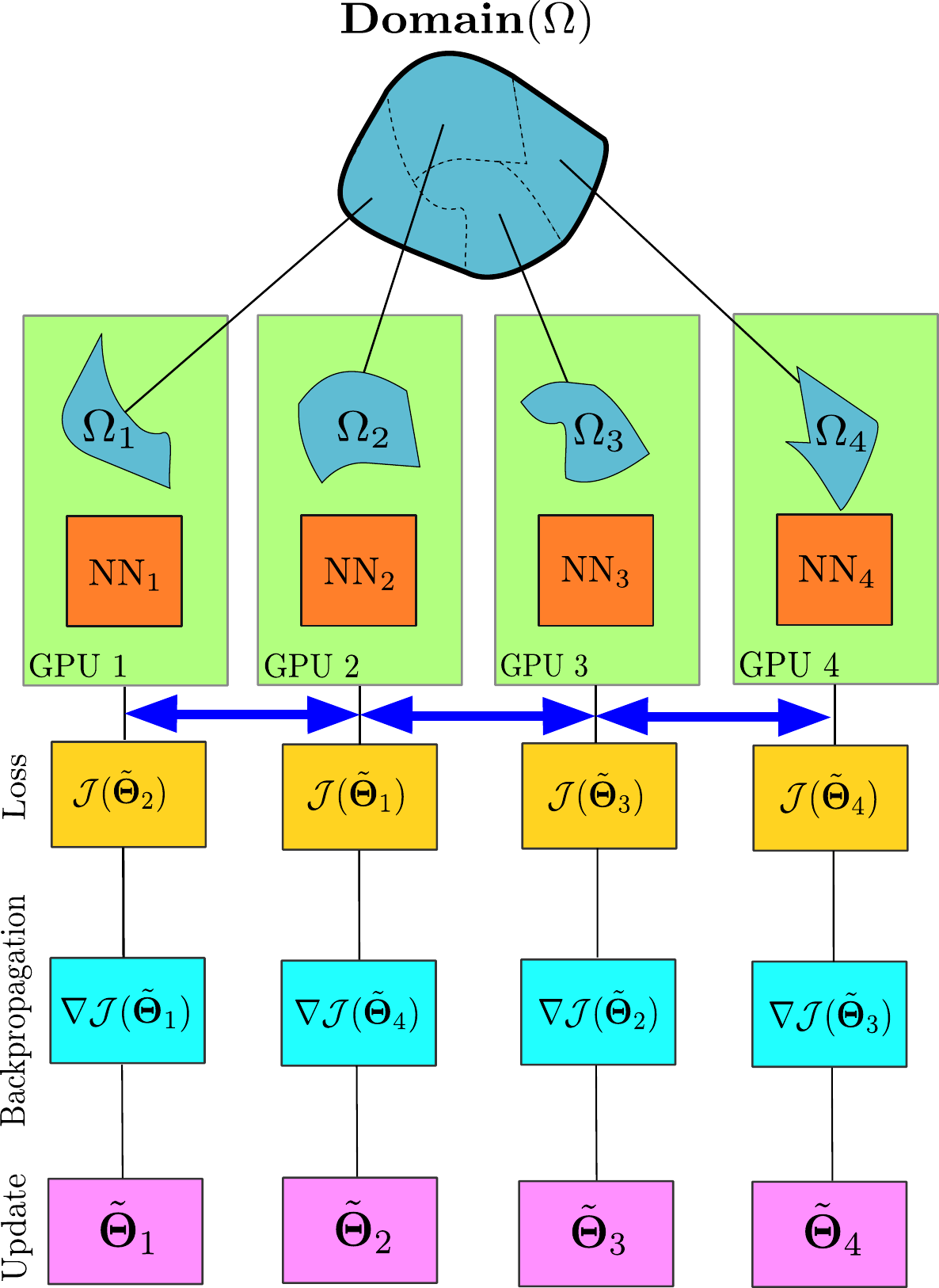}
\caption{\small {\textbf{Parallel PINNs.} Building blocks of distributed cPINN and XPINN methodologies deployed on a heterogeneous computing platform. The domain $\Omega$ is decomposed into several subdomains $\Omega_1,\ldots, \Omega_4$ equal to the number of processors, and individual neural networks (NN$_1, \ldots,$ NN$_4$) are employed in each subdomain, which gives separate loss functions $\mathcal{J}(\tilde{\boldsymbol{\Theta}}_q), q = 1,\ldots,4$ coupled through the interface conditions (shown by the double-headed blue arrow). The trainable parameters are updated by finding the gradient of loss function individually for each network and penalizing the
continuity of interface conditions.}} 
   \label{domain-decomp0}
\end{figure}

\begin{figure}
\centering
\includegraphics[scale=0.35]{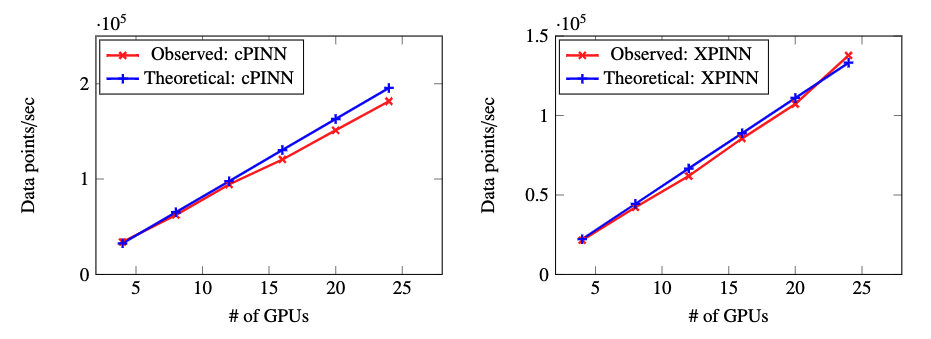}
\caption{\small {Two-dimensional incompressible Navier-Stokes equations: Weak GPU scaling for the distributed cPINN (left) and XPINN (right) algorithms.}} 
\label{fig:weak_scaling}
\end{figure}

 The implementation of cPINN and XPINN on distributed and heterogeneous computing architectures are discussed in detail by Shukla \etal \citep{shukla2021parallel}. The core idea of implementation is based on the domain decomposition approach, where in each subdomain we compute the corresponding solution, and the residuals and solutions at shared edges or planes are passed using message passing interface (MPI) as shown in Fig. \ref{domain-decomp0} to be averaged. Fig. \ref{fig:weak_scaling} shows the weak scaling of cPINN and XPINN for up to 24 GPUs measured on the supercomputer Summit of Oak Ridge  National Laboratory. The weak scaling of cPINN and XPINN shows a very good agreement between observed and theoretical performance. The weak scaling also shows that XPINN has less thoroughput than the cPINN, which could be easily justified as computation of flux typically requires one order less auto-differentiation computation than computation of residuals for XPINN. Shukla \etal \citep{shukla2021parallel} carried out a detailed strong and weak scaling along with communication and computation times required for both  methods. 
 
\begin{figure} [htpb] 
\centering
\includegraphics[trim=0.5cm 0cm 0cm 0cm, clip=true, scale=0.15, angle = 0]{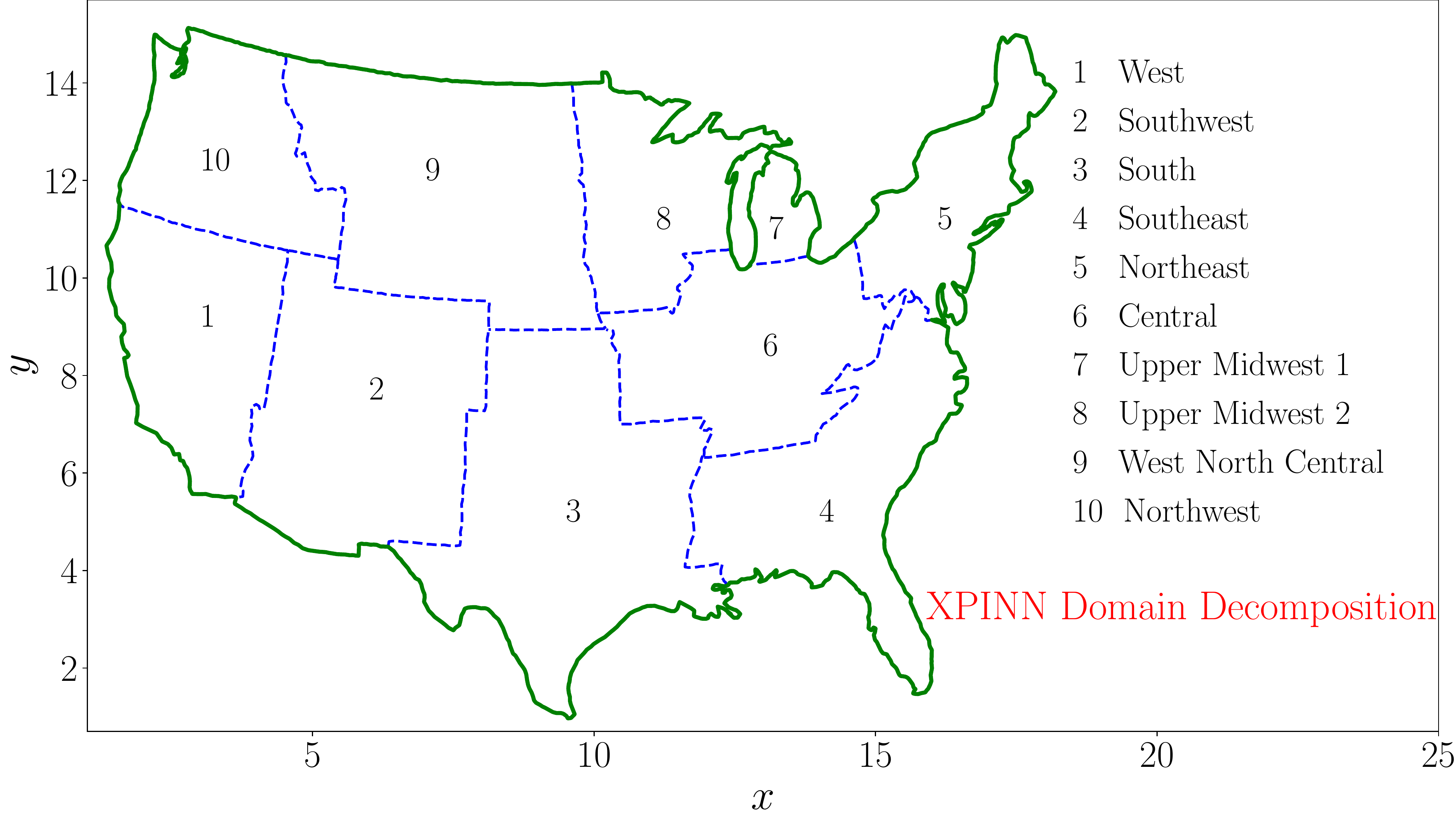}
\includegraphics[trim=0cm 0cm 0cm 0cm, clip=true, scale=0.20, angle = 0]{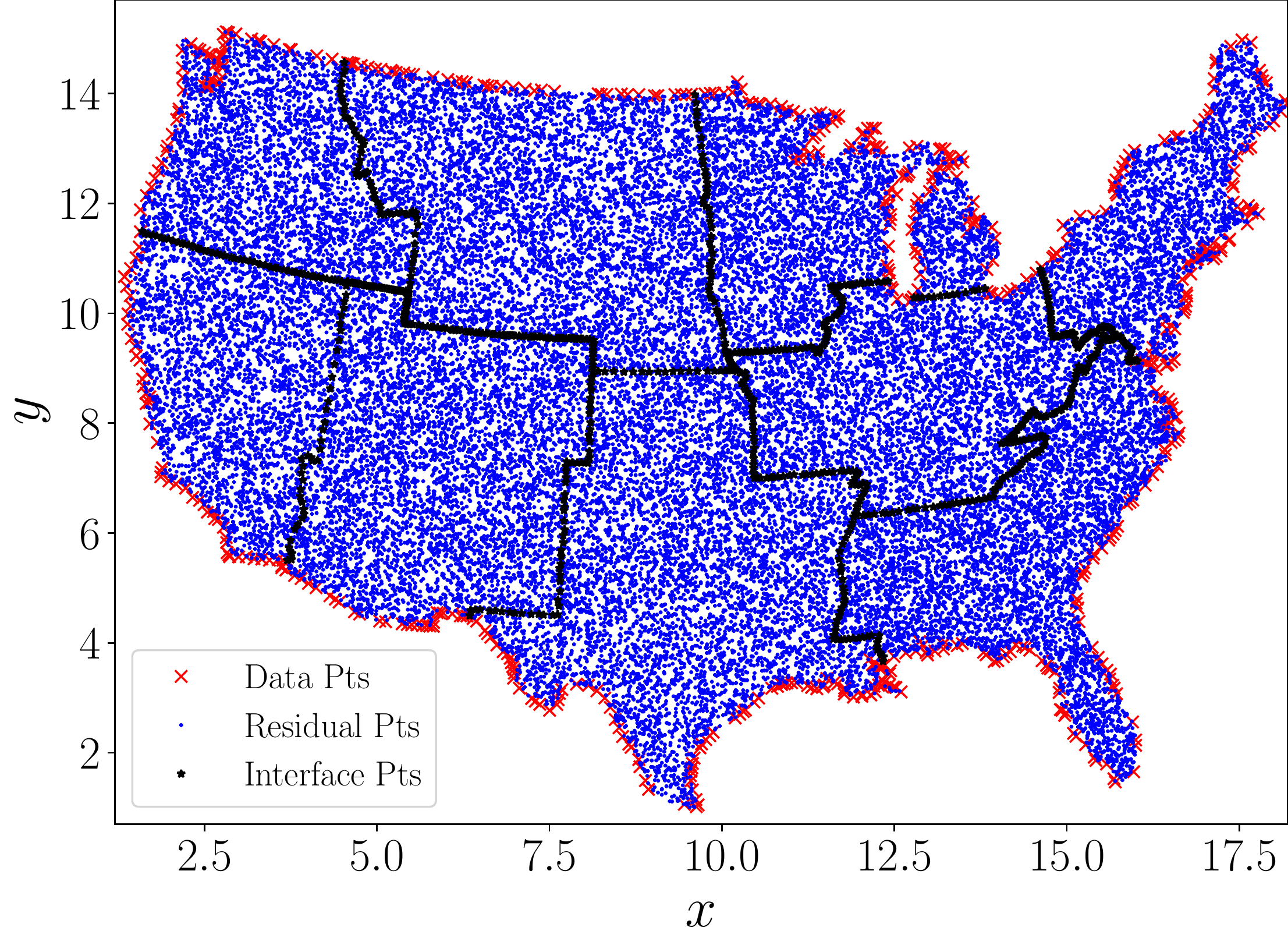}
\caption{\small {Steady-state heat conduction with variable conductivity: Domain decomposition of the US map into 10 regions (left) and the corresponding data, residual, and interface points in these regions (right). }} 
\label{fig:USmapD}
\end{figure}

\begin{figure} [htpb] 
\centering
\includegraphics[trim=0cm 0cm 0cm 0cm, clip=true, scale=0.15, angle = 0]{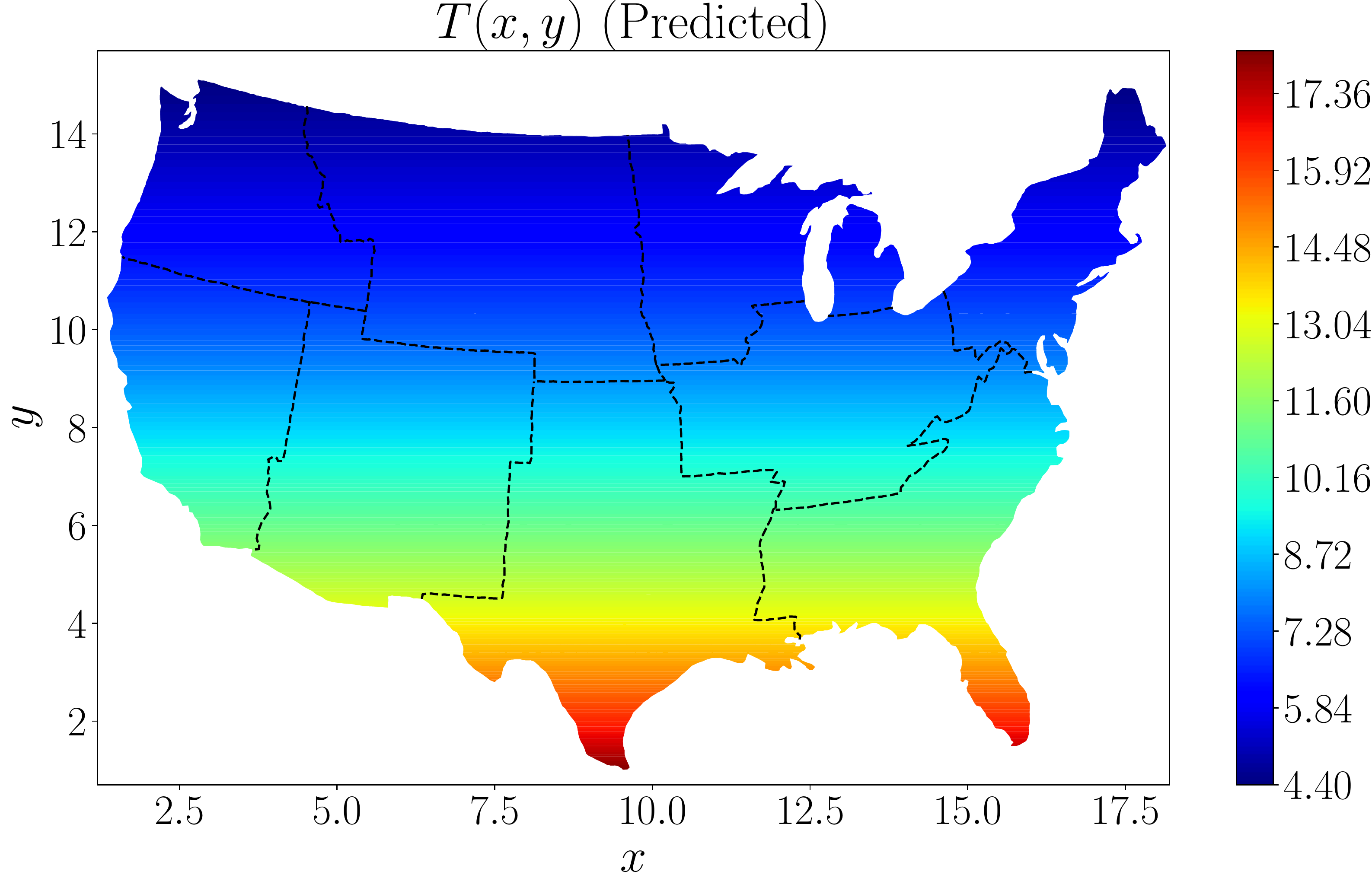}
\includegraphics[trim=0cm 0cm 0cm 0cm, clip=true, scale=0.15, angle = 0]{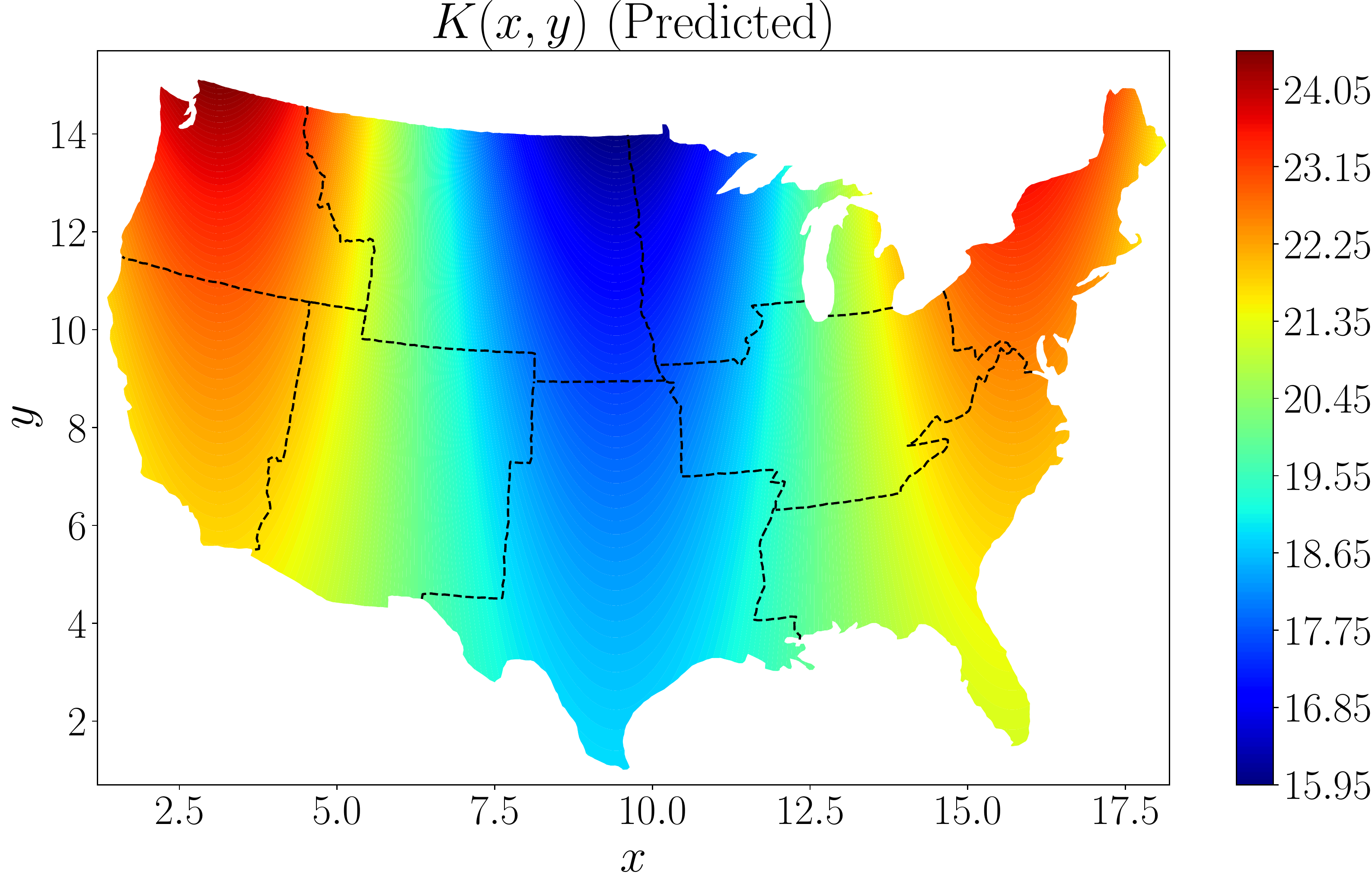}
\includegraphics[trim=0cm 0cm 0cm 0cm, clip=true, scale=0.15, angle = 0]{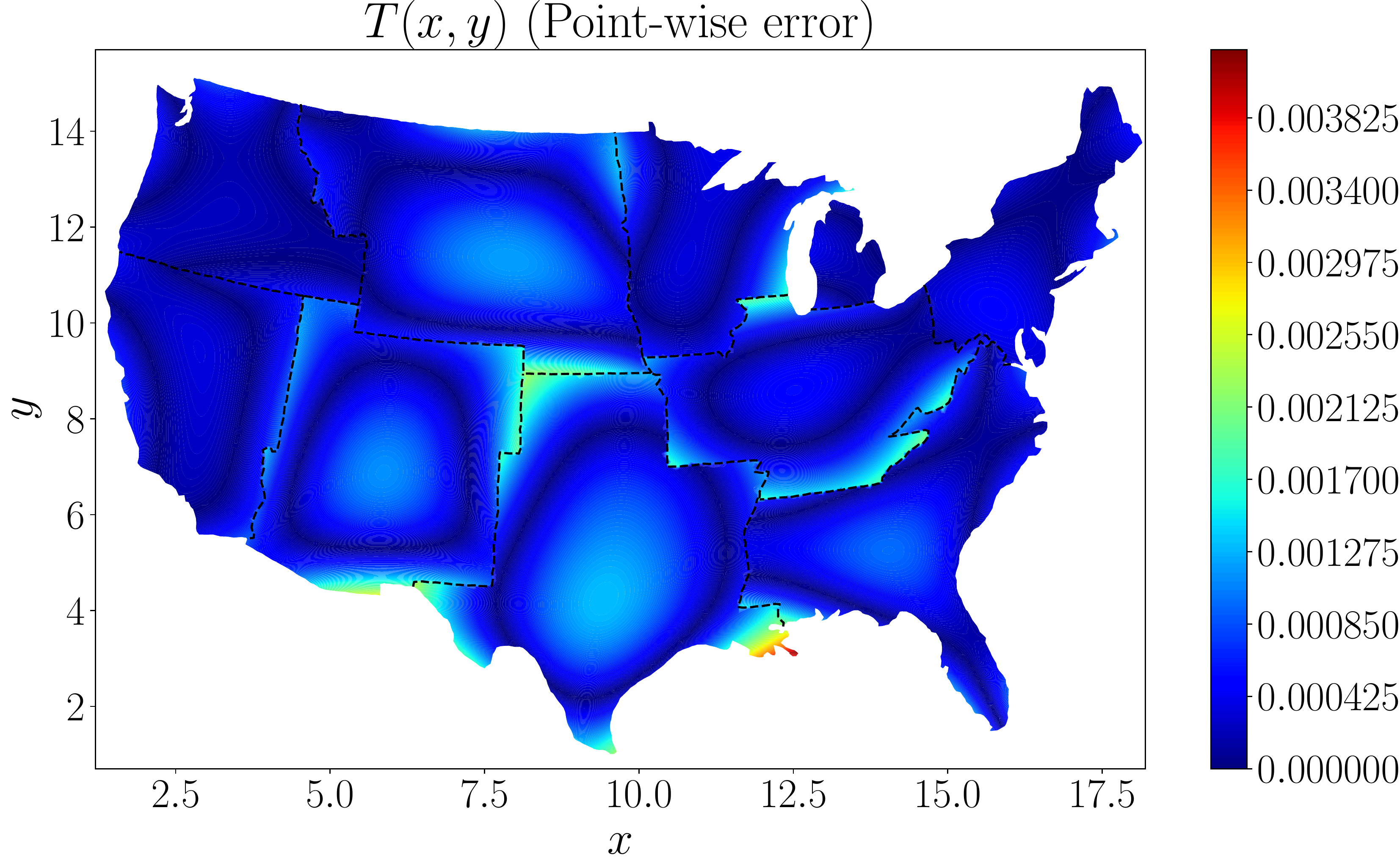}
\includegraphics[trim=0cm 0cm 0cm 0cm, clip=true, scale=0.15, angle = 0]{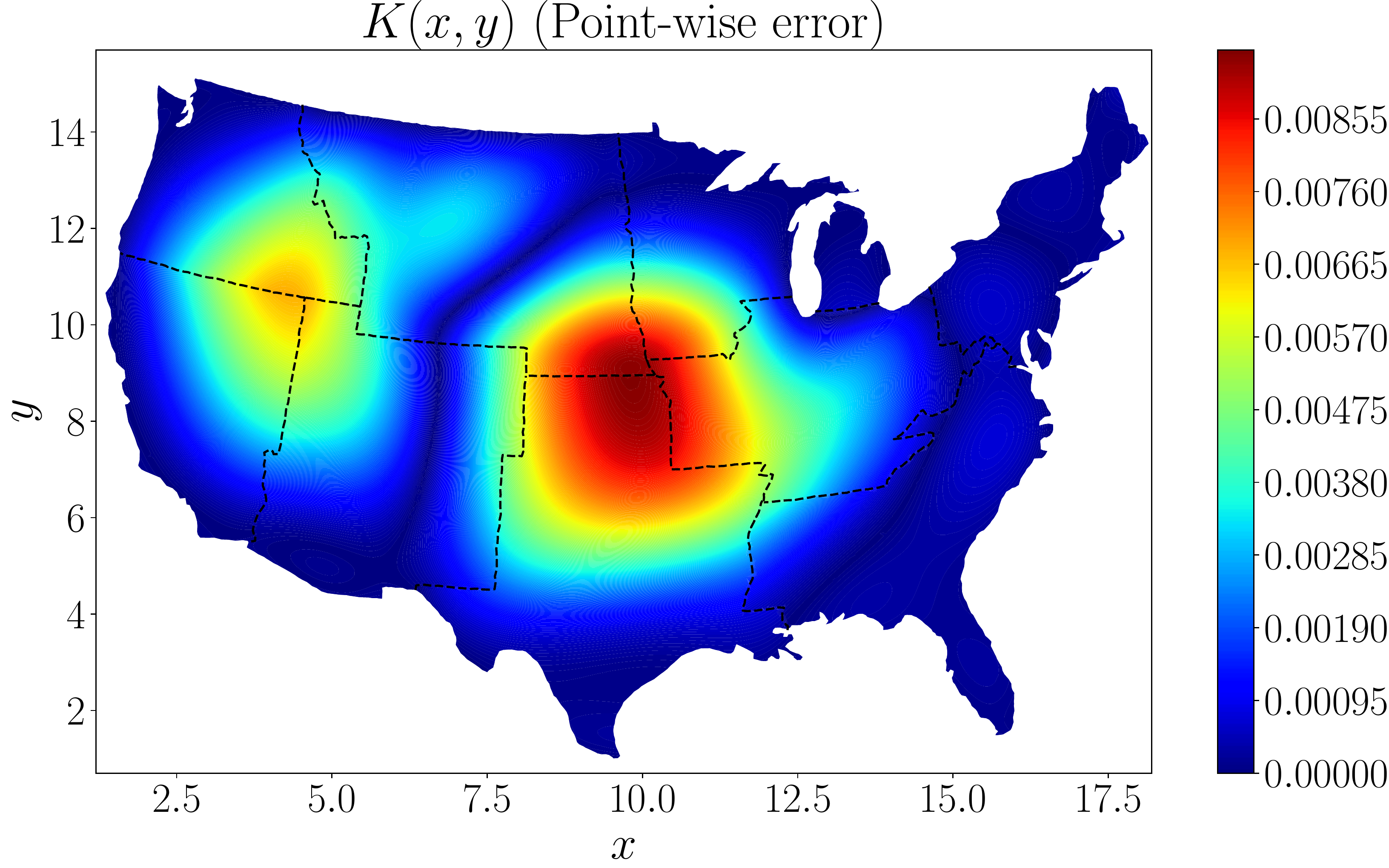}
\caption{\small {Steady-state heat conduction with variable conductivity: The first row shows the contour plots for the predicted temperature $T(x,y)$ and thermal conductivity $K(x,y)$ while the second row shows the corresponding absolute point-wise errors.Adopted from \citep{shukla2021parallel}.}} 
\label{fig:USmap_Sol}
\end{figure}

\begin{figure}
\centering
\includegraphics[trim=0.5cm 0cm 0cm 0cm, clip=true, scale=0.40, angle = 0]{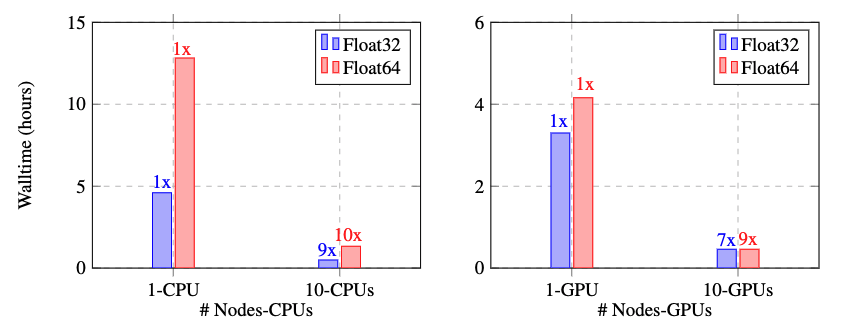}
\caption{\small {Steady-state heat conduction with variable conductivity: Walltime and speedup of parallel XPINN algorithm on CPUs and GPUs implemented for the inverse heat conduction problem in (\ref{heq}); (a) speedup and wall time for the parallel XPINN code on Intel's Xeon(R) Gold 6126 CPU. The speed and wall time is measured for computations performed with single (Float32) and double-precision numbers (Float64); (b) speedup and wall time, measured for single- and double-precision operations, on Nvidia's V100 GPUs.Adopted from \citep{shukla2021parallel}.}}
\label{xpinn_heat_sl}
\end{figure}

\subsection{Inverse problem with parallel PINNs}
To validate the scalability of cPINN and XPINN for a very large inverse problem, we solve the steady-state heat conduction problem over a domain  represented by the USA map.
The governing equation for the steady-state heat conduction problem is written as
\begin{equation}\label{heq}
 \partial_x (K(x,y)T_x) + \partial_y(K(x,y) T_y) = f(x,y)
\end{equation}
with Dirichlet boundary conditions for temperature $T$ and thermal conductivity $K$, obtained from the exact solution. The forcing term $f(x,y)$ is obtained from the exact solution for variable thermal conductivity and temperature, which is assumed to be of the form
\begin{align*}
 T(x,y) &= 20~\text{exp}(-0.1y) 
 \\ K(x,y) &= 20 + \text{exp}(0.1y)~\sin(0.5x).
\end{align*}

Materials whose thermal conductivity is a function of space are the so-called \textit{functionally graded materials}, which are a particular type of composite materials. The material domain is chosen to be a map of the United States divided into 10 subdomains as shown in Fig. \ref{fig:USmapD}. The interfaces between these subdomains are shown by the dashed blue line whereas the domain boundary is shown by the solid green line. The boundary and the training data is available for every subdomain. Unlike cPINN, XPINN can easily handle such complex subdomains with arbitrarily shaped interfaces, thus, providing greater flexibility in terms of domain subdivision. Fig. \ref{fig:USmapD}(right) shows the residual, training data, and interface points in blue dot, red cross, and black asterisk, respectively, over the entire domain.

In this inverse problem, the temperature is assumed to be known in the domain but the variable thermal conductivity is unknown. The aim is to infer the unknown thermal conductivity of the material from a few data points for $K$ available along the boundary line and temperature values available inside the domain.
We employed a single PINN in each subdomain, and the details about the network's architecture are given in the table \ref{Table_USmap}.
\begin{table}[htpb]
\caption{\small {Steady-state heat conduction with variable conductivity: Neural network architecture in each subdomain.}}
\label{Table_USmap}
\begin{center}
\small \begin{tabular}{cccccc} 
\hline 
Subdomain number &  1 & 2&  3 & 4&  5 \\ \hline
$\#$ Residual points & 3000 &4000& 5000 &4000& 3000
 \\
 Adaptive Activation function & $\tanh$ &$\sin$ & $\cos$ & $\tanh$ &$\sin$ 
 \\
 \hline 
 Subdomain number &  6 & 7&  8 & 9&  10
 \\ \hline
 $\#$ Residual points & 4000& 800 &3000& 5000 &4000
 \\
 Adaptive Activation function & $\cos$ & $\tanh$ &$\sin$ & $\cos$ & $\tanh$
 \\
 \hline
 \end{tabular}
\end{center}
\end{table}
In each subdomain, we used 3 hidden-layers with 80 neurons in each layer, and the learning rate is 6e-3, which is fixed for all networks. 

Fig. \ref{fig:USmap_Sol} (top row) shows the predicted values of temperature and thermal conductivity. The absolute point-wise errors are given in the bottom row. 
The XPINN method accurately inferred the thermal conductivity, which shows that XPINN can easily handle highly irregular and non-convex interfaces.

Fig. \ref{xpinn_heat_sl} represents the performance of the parallel XPINN algorithm deployed for solving the inverse heat conduction problem. The performance is measured for the data points given in Table \ref{Table_USmap}. The left sub-figure in Fig. \ref{xpinn_heat_sl} represents the wall time and scaling of the parallel XPINN algorithm on CPUs by computing the solution in each domain using one CPUs. The continuity of solution on the shared boundaries of the subdomain is imposed by passing the solution on each domain boundary to the neighboring subdomains though MPI protocols. Here, each CPU corresponds to a rank mapped by node, and Intel's Xeon(R) Gold 6126 CPUs are used for measurement. First, we measured the wall time of the algorithm on one CPU and considered it as baseline data $(1\text{X})$ for the scaling. Thereafter, we computed the wall time for 10 CPUs, leading to a $(9\text{X})$ and  $(10\text{X})$ for single (32-bit float) and double-precision (64-bit float) computation, respectively. The scaling for double-precision is relatively better as the communication process is shadowed by the computation time due to more time being spent on double-precision arithmetic. However, double-precision-based computation increases wall time by a factor of $2.5$ for single and multiple CPU-based implementations.

Next, we report the walltime and GPU-based implementation. The algorithm is implemented on Nvidia V100 GPUs with 16 GB memories. The hardware architecture is similar to that reported in the example of incompressible Navier-Stokes equations. The right sub-figure of Fig. {\ref{xpinn_heat_sl}} represents the walltime with one and ten GPUs  for single- and double-precision arithmetic. On a single GPU, considered as $(1\text{X})$ model the wall time for double-precision arithmetic 30\% more than that received for single-precision. In the multi-GPU implementation, one GPU is used for each subdomain, and here one rank mapped by the node corresponds to the combination of 1CPU + 1GPU. The walltime for 10-GPUs (10 Nodes or 10 ranks) yields a scaling $(7\text{X})$ and $(9\text{X})$ for single- and double-precision arithmetic. We note that the residual points in each subdomain are not enough to saturate the GPUs and therefore more time is spent on fetching the data to memory and inter-GPU communication. To provide more intuition in the purview of the current implementation, for a typical V100 (16 GB) memory, the single-precision performance is 14 TFLOPs with a memory bandwidth of 900 GB/s and therefore for each byte of transfer (memory to GPU core) 15.6 instruction (FLOPs/ Bandwidth) needs to be issued to occupy the GPUs completely. In the context of the current problem, the partition and therefore the load on GPUs (or CPUs) are static and does not change throughout the computation; thus, subdomain 7, endowed with only 800 residual points, has to wait until all the GPUs (or CPUs) complete their work for the respective subdomain. This also results in slight sub-linear performance. However, the performance presented in Fig. \ref{xpinn_heat_sl} shows a very good linear scaling for CPUs on a heterogeneous (CPU + GPU) architecture. Additionally, the scaling of the presented algorithm on CPUs only architecture concurs with the idea of Daghaghi \etal \citep{daghaghi2021accelerating}, where the authors try to revisit the algorithms used in machine learning to make them faster on CPUs rather than getting fixated with one-dimensional development of specific hardware to run the matrix multiplication faster. In this test case, the partition of the domain is performed  by manually choosing the interface points. This was done to show the efficacy of the XPINN algorithm for non-convex or irregular subdomains. In such complex subdomains, the distribution system often faces the problem of load imbalance, which can seriously degrade the performance of the system. A more efficient approach could be utilized to decompose the domain such that the subdomains are packed optimally for inter-node or inter-process communication. A suitable point cloud \citep{Rusu_ICRA2011_PCL}  or K-way partition \citep{METIS} based approach will result in further optimizing the communication.

\section{Graph Neural Networks}
\label{sec:PIGNs}
In contrast to more traditional ML data handled in image recognition or natural language processing, scientific data is fundamentally unstructured, requiring processing of polygonal finite element meshes, LIDAR point clouds, Lagrangian drifters, and other data formats involving scattered collections of differential forms. This poses both challenges and opportunities. Convolutional networks serve as a foundation for image processing, and are able to employ weight-sharing by exploiting the Cartesian structure of pixels, but cannot be applied in unstructured settings where stencils vary in size and shape from node to node. Unstructured data do, however, encode nontrivial topological information regarding connectivity, allowing application of ideas from topological data analysis and combinatorial Hodge theory \citep{wasserman2018topological,bubenik2015statistical,jiang2011statistical,carlsson2009topology}. 
In contrast, Graph Neural Networks (GNNs) form a class of deep neural network methods designed to handle unstructured data. Graphs serve as a flexible topological structure, which may be used for learning useful latent ``graph-level'', ``node-level'' or ``edge-level'' representations of data for diverse graph prediction tasks and analytics \citep{monti2017geometric}. 

This unstructured nature enables GNNs to naturally handle a wide range of graph analytics problems (i.e., node classification, link prediction, data visualization, graph clustering community detection, anomaly detection) and have been applied effectively across a diverse range of domains, e.g., protein structure prediction~\citep{jumper2021highly}, untangling the mathematics of knots~\citep{davies2021advancing}, brain networks~\citep{rosenthal2018mapping,xu2020new,xu2021graph} in brain imaging, molecular networks~\citep{liu2019chemi} in drug discovery, protein-protein interaction networks~\citep{kashyap2018protein} in genetics, social networks~\citep{wang2019user} in social media, bank-asset networks~\citep{zhou2019asset} in finance, and publication networks~\citep{west2016recommendation} in scientific collaborations. 

In this section we consider the task of fitting dynamics to data defined on a set of nodes $\mathcal{N}$, often associated with an embedding as vertices $\mathcal{V} \subset \mathbb{R}^d$. We will consider the graph $\mathcal{G}=(\mathcal{V}, \mathcal{E})$, where  $\mathcal{E}$ represents the edge set denoted by $\mathcal{E} \subseteq\left(\begin{array}{l}\mathcal{V} \\ 2\end{array}\right)$. We will also consider higher-order k-cliques consisting of ordered k-tuples of nodes, where for example, an edge $e \in \mathcal{E}$ corresponds to a 2-clique. We associate scalar values $\bm{x}$ with nodes, edges, or both, and aim to either learn dynamics ($\dot{\bm{x}} = L(\bm{x};\theta)$) or boundary value problems ($L(\bm{x};\theta) = f$) for learnable parameters $\theta$.

We present three approaches: 1) methods which aim to learn operators via finite difference-like stencils, 2) graph neural networks which represent operators by learning update and aggregation maps, and 3) graph calculus methods, which relate learnable operators to graph analogs of the familiar vector calculus div/grad/curl. First, we survey prevailing strategies and architectures for traditional GNNs.

\subsection{Basics of GNNs}

Broadly, GNNs may be classified into three major categories: 1) Spectral GNNs, which locally aggregate connected node information by solving eigenproblems in the \textit{spectral domain}. 2) Spatial GNNs, which perform graph convolution via aggregating node neighbor information from the first-hop neighbors in the \textit{spatial graph domain}. 3) Graph Attention Networks (GATs), which leverage the \textit{self-attention} mechanism for hidden feature aggregation. We recall first some mathematical definitions before presenting each in turn.

\paragraph{Notation and mathematical preliminaries} Consider a node $v_{i} \in \mathcal{V}$ and $e_{i, j}=\left\{v_{i}, v_{j}\right\} \in \mathcal{E}$ indicating either a directed or undirected edge between nodes $v_i$ and $v_j$. By $j \sim i$, we denote neighbors $j$ with corresponding edges $(i,j)$ coincident upon node $i$.  $\mathbf{A} \in \mathbb{R}^{N \times N}$ corresponds to the adjacency matrix of $\mathcal{G}$. When $\mathcal{G}$ is binary (or ``unweighted''), $\mathbf{A}_{i, j} = 1$ if there is a edge between nodes $v_i$ and $v_j$, else 0. If $\mathcal{G}$ is a weighted graph, $\mathbf{A}_{i, j} = w_{ij}$ when there is an edge between nodes $v_i$ and $v_j$. A graph is \textit{homogeneous} if all nodes and edges are of a single type, and vice versa, the graph is \textit{heterogeneous} if it consists of multiple types of nodes and edges. Additionally, each node may be associated with features (or attributes) $\mathbf{x} \in \mathbb{R}^{N \times F}$  referring to the node attribute matrix of $\mathcal{G}$. 
The entry $x_{i} \in \mathbf{x}$ corresponds to the node feature vector of node $v_i$. The graph Laplacian matrix is an $N \times N$ matrix defined as $\mathbf{L}=\mathbf{D}-\mathbf{A}$, where $\mathbf{D}$ is a $N \times N$ diagonal node degree matrix whose $i$-th element is the degree of the node $i$, i.e., $\mathbf{D}(i,i) = \sum_{j = 1}^{N} A_{i,j}$.

\paragraph{Spectral Methods.} 
The main idea of spectral GNN approaches is to use eigen-decompositions of the graph Laplacian matrix $\mathbf{L}$ to generalize spatial convolutions to graph structured data. This allows simultaneous access to information over short and long time scales \citep{bruna2013spectral}. Spectral graph convolutions are defined in the spectral domain based on the graph Fourier transform.

Specifically, the normalized self-adjoint positive-semidefinite graph Laplacian matrix $\mathbf{L}$ describes diffusion over a graph~\citep{chung1997spectral}, and can be factorized by $\mathbf{L} = U \Lambda U^{\top}$. $U$ is $N\times N$ matrix whose $l$-th column is the eigenvector ($u_l$) of $\mathbf{L}$; $\Lambda$ is a diagonal matrix whose diagonal elements $\lambda_l = \Lambda(l,l)$ ($l \in \mathcal{R}^N$) are the corresponding eigenvalues. We can apply the operation $U^{\top} \phi$ to project data $\phi$ from graph nodes into the spectral domain, where features can be decomposed into a sum of orthogonal graphlets or motifs (i.e., eigenvectors) that make up $\mathcal{G}$. Typically, eigenvectors with the smallest eigenvalues may be used to distinguish vertices that will be slowest to share information under diffusion or message passing. 
Projecting vertex features onto these eigenvectors and processing the projected graph signal is equivalent to augmenting message passing with a nonlinear low-pass filter. 

Recently, many spectral GNN variants have been proposed, e.g., \citep{bruna2013spectral}, \citep{kipf2016semi}. The main workflow of spectral GNNs follows four main steps: 1) transform the graph to the spectral domain using the graph Laplacian eigenfunctions (see Eq.~\ref{eq:graph_FFT}); 2) perform the same transformation on the graph convolutional filters; 3) apply convolutions in the spectral domain; 4) transform from the spectral domain back to the original graph domain (see Eq.~\ref{eq:inverse_transform}). 
\begin{equation}
\label{eq:graph_FFT}
  \hat{\mathbf{x}}\left(\lambda_{l}\right)=\left\langle\mathbf{x}, \mathbf{u}_{l}\right\rangle=\sum_{i=1}^{N} \mathbf{x}(i) \mathbf{u}_{l}^{T}(i)  
\end{equation}
\begin{equation}
\label{eq:inverse_transform}
\mathbf{x}(i)=\sum_{l=1}^{N} \hat{\mathbf{x}}\left(\lambda_{l}\right) \mathbf{u}_{l}(i).
\end{equation}
The detailed spectral graph convolution operation is described in Eq.~\ref{eq:spectral_graph_conv}, where $U$ is the eigenvector matrix, $X^{(k)}$ represents the node feature map at the $k$-th layer, and $W^{(k)}$ denotes the learnable spectral graph convolutional filters at the $k$-th layer.
\begin{equation}
X^{(k+1)}=U\left(U^{T} X^{(k)} \odot U^{T} W^{(k)}\right).
\label{eq:spectral_graph_conv}
\end{equation}

However, the aforementioned spectral GNN methods share a few significant weaknesses: the Laplacian eigenfunctions are inconsistent across different domains, and therefore the method generalizes poorly across different geometries. Moreover, the spectral convolution filter is applied for the whole graph without considering the local graph structure property. In addition, the graph Fourier transform has high cost in computation. To solve the locality and efficiency problem, ChebNets~\citep{defferrard2016convolutional} employs the Chebyshev polynomial basis to represent spectral convolution filter instead of graph Laplacian eigenvectors. 

\paragraph{Spatial Methods.} 

Spatial methods involve a form of learnable message-passing that propagates information over the graph via a local aggregation (or diffusion) process. Graph convolutions are defined directly on the graph topology in spatial methods~\citep{atwood2016diffusion, niepert2016learning, gilmer2017neural}. Spatial GCNs directly ignore the ``edge attributes'' as well as the crucial ``messages'' sent by nodes along edges in the graph, however, the Message Passing Neural Networks (MPNNs) can effectively make use of this information to update node states. MPNN consists of three main phases: \textit{message passing, readout and classification}. Message passing consists of: 1) A node feature transformation based on some sort of projection. 2) Node feature aggregation with a permutation-invariant function, e.g., averaging, sum or concatenation. 3) A node feature update via the current states and representations aggregated from each node's neighborhood. Specifically, the message passing runs for $T$ time steps, and includes two key components: message functions $M_t$ and node update functions $U_{t}$. The hidden states $h_{i}^{t}$ at each node in the graph are updated based on Eq.~\ref{eq:update_hidden_states} with the current state ($h_{i}^{t}$) and the aggregated messages $m_{i}^{t+1}$, which are computed from neighboring nodes' feature ($h_{j}^{t}$) and edge features ($e_{ij}$) according to Eq.~\ref{eq:message_passing}.
\begin{equation}
\label{eq:update_hidden_states}
h_{i}^{t+1}=U_{t}\left(h_{i}^{t}, m_{i}^{t+1}\right),
\end{equation}
\begin{equation}
\label{eq:message_passing}
m_{i}^{t+1}=\sum_{j \in \mathcal{N}_i} M_{t}\left(h_{i}^{t}, h_{j}^{t}, e_{ij}\right), 
\end{equation}
where $\mathcal{N}_i$ denotes the neighbors of node $i$. The readout phase computes a feature vector for the whole graph using some readout function $R$ according to
$$
\hat{y}=R\left(\left\{h_{i}^{T} \mid i \in \mathcal{G}\right\}\right).
$$
The message functions $M_{t}$, node update functions $U_{t}$ and readout function ($R$) are all learned differentiable functions. Additionally, $R$ must be invariant to node permutation (i.e., graph isomorphism) to guarantee identical output for equivalent graphs.


\paragraph{Graph Attention Networks (GATs).}
In spectral and spatial GNN models, the message aggregation operations from nodes’ neighborhoods are mostly guided by the graph structure, which weights contributions from neighboring nodes equally. In contrast, GATs~\citep{velivckovic2017graph} assign during aggregation a learnable propagation weight via a ``self-attention'' function ($\alpha$), defined in Eq.~\ref{eq:alpha_attention_func}. Specifically,
given the node pair feature vectors ($\vec{h}_{i}$ and $\vec{h}_{j}$), the attention coefficients ($e_{i,j}$) between nodes $i$ and $j$ can be computed by using a self-attention mechanism $a: \mathbb{R}^{F^{\prime}} \times \mathbb{R}^{F^{\prime}} \rightarrow \mathbb{R}$ ($F^{\prime}$ is the node feature size), which is usually implemented by a single-layer feed-forward neural network. To this end, the attention mechanism can help identify the importance of interactions between each node neighbor. Here, $c_{i,j}$  reflects the importance of node $j$'s feature to node $i$. For better comparability across all nodes in the graph, the normalized attention coefficients ($\alpha_{i,j}$) between node pair $(i,j)$ can be defined as a softmax function (see Eq.~\ref{eq:alpha_attention_func}), where $\mathcal{N}_{i}$ denote the neigborhood of node $i$, i.e., first-order neighbors.
\begin{equation}
\label{eq:alpha_attention_func}
    \alpha_{i j}=\frac{\exp \left(c_{i j}\right)}{\sum_{k \in \mathcal{N}_{i}} \exp \left(c_{i k}\right)}, c_{i j}=a\left(h_{i}, h_{j}\right).
\end{equation}

In practice, a multi-head attention mechanism has been widely used in GATs which can stabilize the process by using $K$ independent attention mechanisms to achieve the transformation. The corresponding learned node feature for node $i$ with multi-head attention mechanism~\citep{vaswani2017attention} is shown in Eq.~\ref{eq:GAT}. 
\begin{equation}
\label{eq:GAT}
    h_{i}^{\prime}=\|_{k=1}^{K} \sigma\left(\sum_{j \in \mathcal{N}_{i}} \alpha_{i j}^{k} \mathbf{W}^{k} h_{j}\right),
\end{equation}
where $\alpha_{i j}^{k}$ denotes the attention coefficients derived by the $k$-th attention head; $\mathbf{W}^{k}$ is the weight matrix specifying the linear transformation of the $k$-th attention head; $\sigma$ is the nonlinear activation function; $\|$ represents ultimate concatenation, sum or average operation for the features learned by $K$ independent attention heads.

In summary, GCNs are a special case of GATs which use the spectral normalized graph adjacency matrix to serve the role of the attention function. GATs are a special case of MPNNs with hidden feature aggregation by using self-attention mechanism as message passing rule. All methods share common inductive bias and node/edge permutation invariance properties. Lastly, there are also some works (DyRep~\citep{trivedi2019dyrep}, EvolveGCN~\citep{pareja2020evolvegcn}, TGN~\citep{rossi2020temporal}, DynG2G~\citep{xu2021dyng2g}, etc.) for {\em dynamic} graph structure data inference (climate network, traffic network, etc.).


\subsection{Learning stencil operators}
In this section, we consider learning operators expressed via a discrete stencil
\begin{equation}
    L(\bm{x};\theta) = \sum_{i \in \mathcal{V}} \theta_i \bm{x}_i,
\end{equation}
where the learnable parameters $\theta_i$ correspond to stencil coefficients. An attractive feature of this simple setup is that differential operators may be encoded as a finite difference stencil. However, unlike traditional finite difference methods, often we consider a fixed size graph and do not consider a sequence of refined grids which converge to some continuum operator. Therefore, we are more interested in recovering an effective discrete operator in the traditional finite difference sense.

\begin{figure}[!ht]
\centering
\includegraphics[scale=0.3]{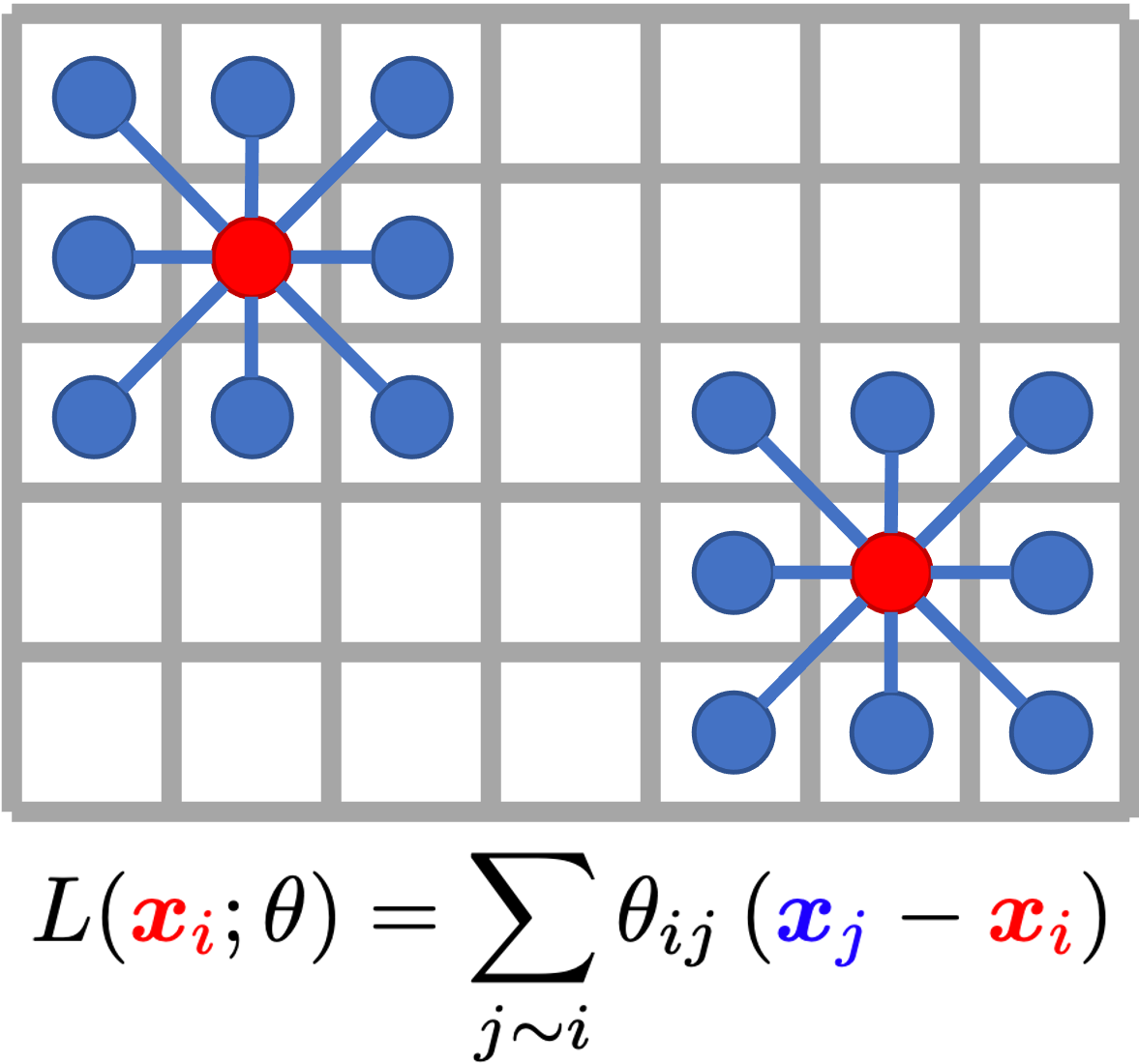}\quad
\includegraphics[scale=0.3]{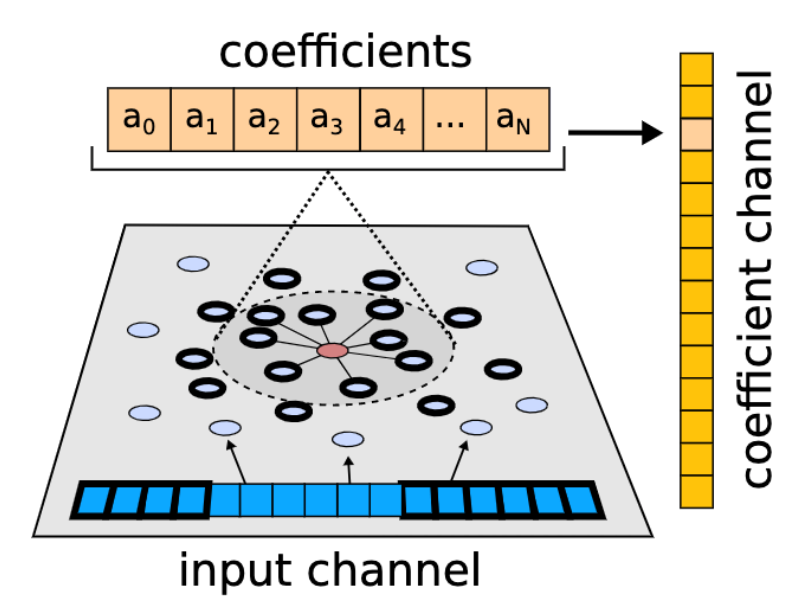}\\
\includegraphics[scale=0.3]{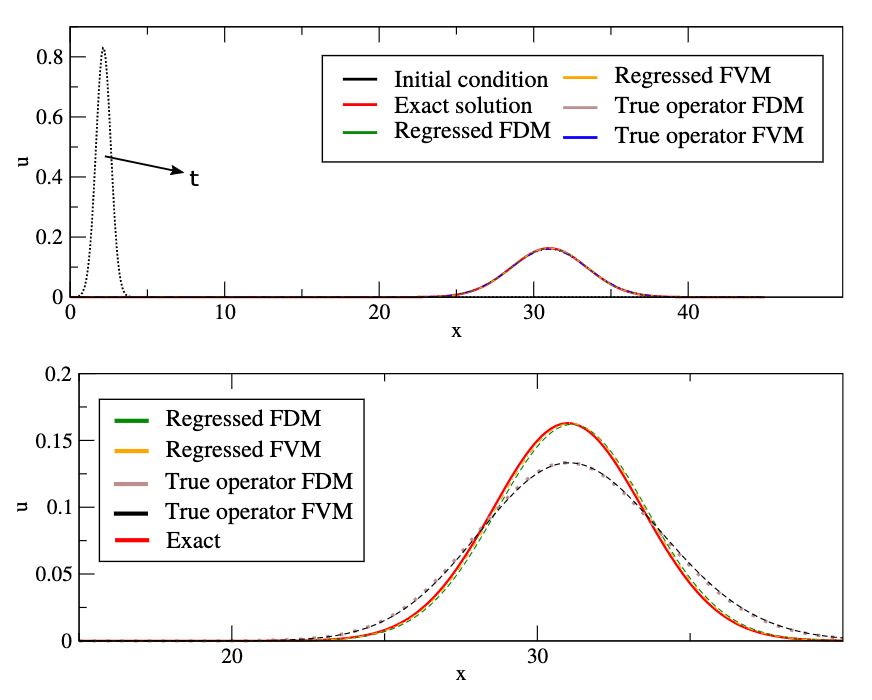}
\includegraphics[scale=0.3]{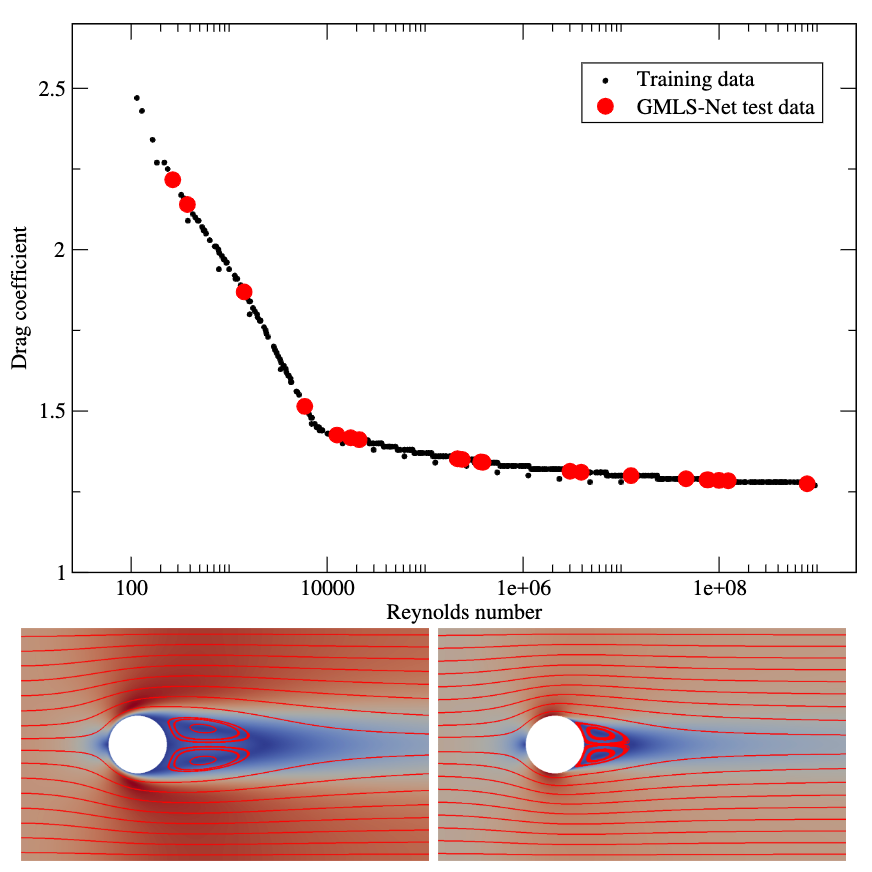}
\caption{\small {\textbf{Learning operators from stencils.} \textit{From left to right, top to bottom:} (1) On a Cartesian grid of data, CNNs may employ weight-sharing to fit finite-difference operators to data \citep{bar2019learning}. (2) On unstructured data, similar weight-sharing may be achieved by lifting data first to a space of polynomial coefficients and learning how an operator acts on polynomials. (3) Learning operators as difference stencils allows learning of higher-order corrections. Shown here at a CFL condition of 10, a stencil learned from an analytic solution to the advection-diffusion problem does not exhibit the numerical dissipation of a traditional finite difference/volume discretization of the PDE. (4) Beyond learning physics, these frameworks can be used for supervised learning tasks on unstructured scientific data. Shown here, the drag force acting on a cylinder is regressed from nodal velocities on an unstructured finite volume mesh. Figures adopted from \citep{trask2019gmls}}} 
\label{fig:stencils}
\end{figure}

In \citep{bar2019learning}, the authors repurpose a convolutional network to this end. We note that this corresponds to a graph in which $\mathcal{V}$ consists of a Cartesian grid of nodes, while $\mathcal{E} = \{ (i,j)\,|], |x_i - x_j|_1 < \epsilon \}$ consists of neighbor nodes in an $\epsilon$ ball in the taxicab norm; see also Fig. \ref{fig:stencils}. While simple, this work is extremely interpretable, and the authors demonstrate an ability to learn stencils for Burgers, Korteweg–de Vries, and Kuramoto–Sivashinsky equations with improved error compared to a naive application of the finite difference method. We recall from introductory finite difference methods (\citep{gustafsson1995time}) that schemes such as the Lax-Wendroff or Macormack schemes introduce higher-order dissipation or dispersion into a discrete operator to achieve improved accuracy or stability. A simple Taylor series analysis reveals such schemes incorporate higher-order differential correction terms which depend on the spatiotemporal resolution, hence learning finite difference stencils may also learn similar higher-order corrections from data.

The requirement of a Cartesian node set is, however, restrictive and precludes application to complex geometries. In \citep{trask2019gmls} unstructured data is handled by finding a polynomial fit to data at each node of an unstructure $\epsilon-$ball graph and then providing the resulting coefficient weights to a multilayer perceptron.
\begin{equation}
    \min_{\mathbf{c}_i}\sum_{j ~ i \in \mathcal{V}} (u_j - \mathbf{c}_i^\intercal P(x_j) )^2 W(|x_i - x_j|) \qquad L(\bm{x};\theta) = \mathcal{N}(\mathbf{c}_i(\bm{x});\theta)
\end{equation}
Here, $P$ denotes a vector of polynomial basis functions, $W$ denotes a radial kernel with support $B_\epsilon(x_i)$, and $\mathcal{N}$ denotes a multilayer perceptron with weights and biases $\theta$ taking the optimal polynomial reconstruction as input. In \citep{trask2019gmls} this architecture is shown to be effective for both learning physical operators on unstructured data as well as offering an alternative to spectral GNNs for supervised learning on structured and unstructured data; see Fig. \ref{fig:stencils}. This framework builds upon the generalized moving least squares (GMLS) theory, which has traditionally been used to develop finite difference discretizations on unstructured points \citep{mirzaei2012generalized}. In relation to spectral GNNs, replacing a global spectral approximation with a localized polynomial approximation, which may be trivially accelerated with GPUs \citep{compadre_toolkit} is attractive from both performance and approximation perspectives.

\subsection{Physics-Informed Graph Networks (PIGNs)}

The stencil formulation of the previous section prompts a natural interpretation of GNNs in the context of finite difference methods. In staggered finite difference schemes for PDEs, one uses a stencil to compute a conservation balance (e.g., a divergence operator for conservation of fluxes in marker-and-cell schemes for transport \citep{harlow1965numerical}, or a curl operator for circulations in the Yee scheme for electromagnetism \citep{yee1966numerical}) from fluxes/circulations computed on a dual grid. The aggregation step of GNNs admits interpretation as a graph divergence, with message passing corresponding to learning fluxes/circulations.
In \citep{kumar2021grade} the authors apply this viewpoint, using an attention mechanism \citep{velickovic2017graph} to learn messages from data. They demonstrate accurate prediction of solution operators for the viscous Burgers equations in one- and two-dimensions.

In \citep{pfaff2020learning}, the authors consider a similar message-passing GNN where the graph is taken to be a traditional simulation mesh. By augmenting the network with edge features encoding the displacement of the domain from a reference configuration, they demonstrate how GNNs may be used to train surrogates of physics-based models from high-fidelity simulations on unstructured meshes in either an Eulerian or Lagrangian configuration. They demonstrate speed ups of $10-100\times$ for cloth dynamics, elasticity, and compressible flows and attribute good generalization beyond parameters used in training to invariances preserved by the GNN architecture. Also, worth noting is favorable comparison against convolutional methods, where message-passing GNNs provide improved accuracy.

Whereas this is a useful perspective of learning operators to mimic integral balance equations corresponding to PDE models of physics, for design of "systems-of-systems" one may use a graph to represent individual subsystems as nodes and their interactions via edges. This often corresponds to characterizing a control volume consisting of an entire subsystem, rather than an infinitesimal volume. In \citep{hall2021ginns}, the authors develop graph informed neural networks (GINNs) which employ a probabilistic graph model to perform system scale design of supercapacitor dynamics and a Langmuir adsorption model. While the Bayesian network architecture is distinct from the GNN architectures discussed in the previous section, their work provides an example of how graph architectures may be used to replace computationally intensive bottlenecks in system dynamics. In other frameworks aligning more closely with GNNs, similar strategies may be used to learn and design, e.g., system-level circuit design with constituent data-driven device physics \citep{gao2020physics} or heat transfer of industrial buildings \citep{drgovna2021physics}.

\subsection{Graph exterior calculus}
\begin{figure}
\centering
\includegraphics[scale=0.6]{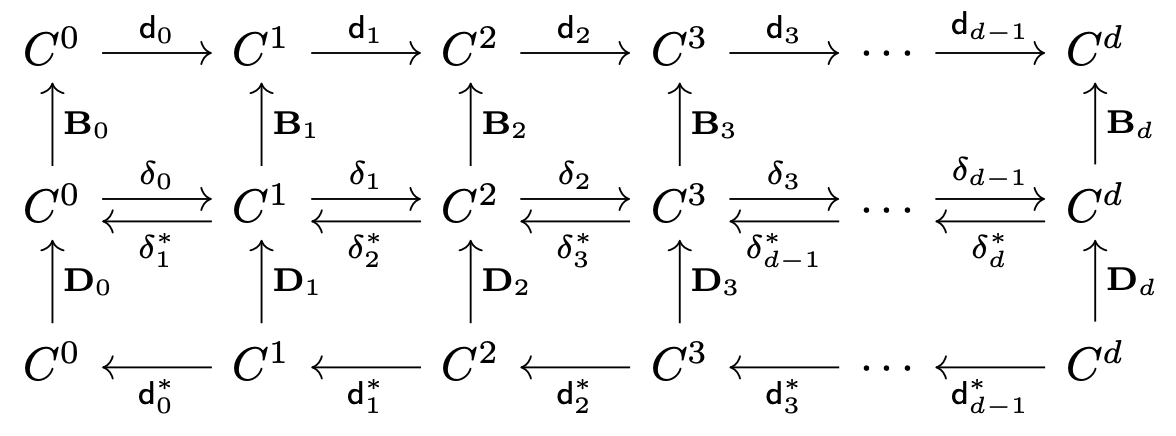}\\
\includegraphics[scale=0.6]{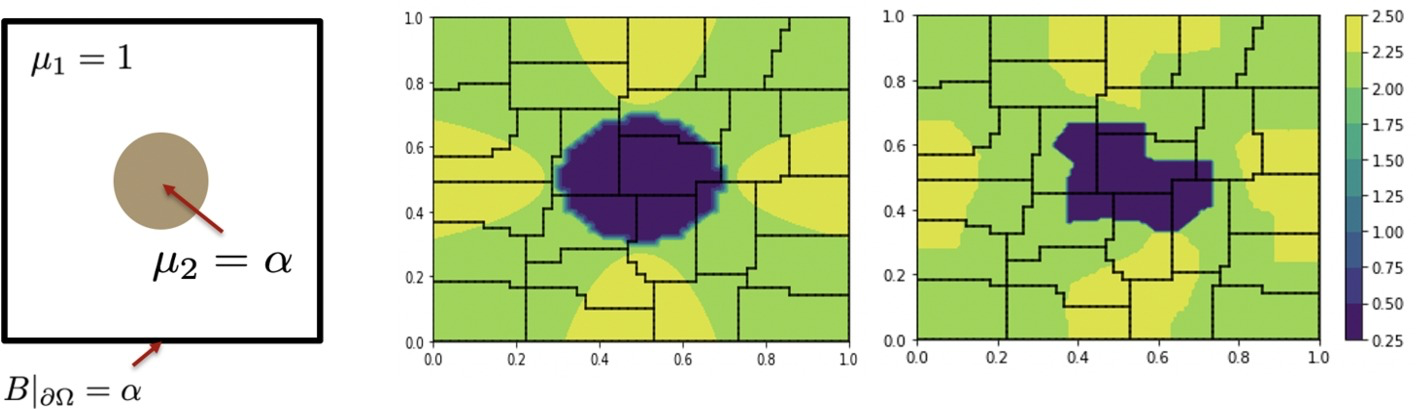}
\includegraphics[scale=0.4]{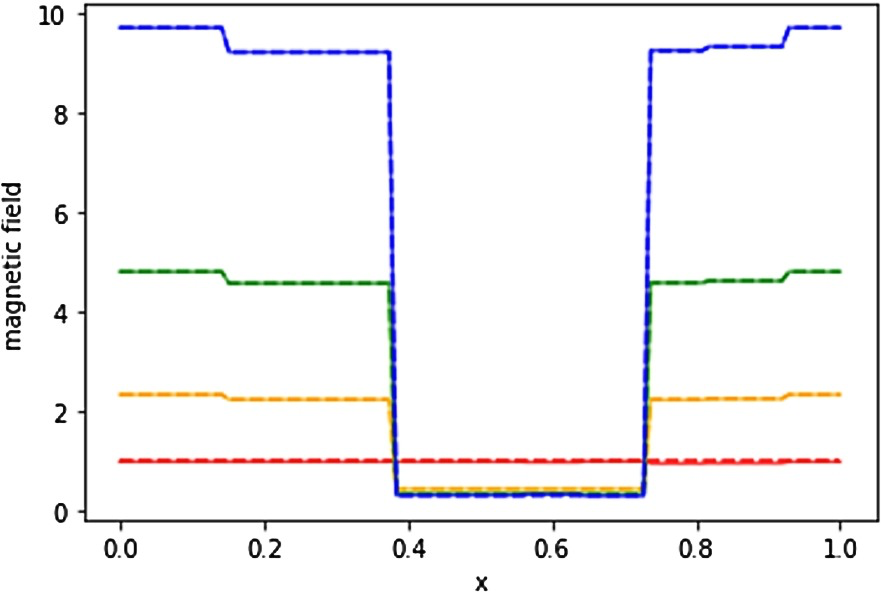}
\caption{\small {\textbf{Graph exterior calculus for physics-informed GNNs.} \textit{Top:} The graph calculus div/grad/curl ($\delta_k$) and their adjoint ($\delta_k^*$) may be augmented with machine learnable metric information ($\bm{B}_k,\bm{D}_k$) to obtain a \textit{de Rham complex}, which preserves algebraic structure related to stability and physical invariances. \textit{Bottom, left to right:} To construct a surrogate for a heterogeneous magnetostatics inclusion problem parameterized by the jump in permittivity $\alpha$ (1), we train a GNN to reproduce solution moments of a high-fidelity solution (2) on a coarse grid (3) while preserving conservation structure. A plot of the magnetic field across $y=0.5$ reveals the predicted solution (dashed) matches the data (solid) to machine precision for a range of jumps ($\alpha = 1,2,4,8$ in red/yellow/green/blue, respectively), demonstrating the scheme handles jumps in material properties for problems with nontrivial null-spaces, similar to mimetic PDE discretizations. Figure adapted from \citep{trask2022enforcing}.}} 
\label{fig:ddec}
\end{figure}
Exterior calculus has been used as a tool for designing mimetic discretizations of PDEs, which preserve structure and properties from continuum solutions such as conservation, maximum principles, nonoscillatory solutions near material discontinuities, and preservation of invariants \citep{arnold2007compatible}. These methods exploit connections to the generalized Stokes theorem to construct a data-driven $\textit{de Rham complex}$ - a commuting diagram (see Figure \ref{fig:ddec}) which allows a unified analysis of discretizations spanning mixed finite elements, staggered finite difference schemes, and mimetic difference methods \citep{bochev2006principles}. Graphs admit a similar \textit{combinatorial de Rham complex} based upon graph extensions of the familiar vector calculus div/grad/curl, which allows rigorous analysis of Hodge Laplacians generalizing the previously discussed graph Laplacian. We discuss several recent works which use this as a basis for machine learning.

Given a collection of $k-$cliques with associated scalar values, one may define a graph calculus. Denote the scalar value associated with a $k-$clique $f_{{i_1},...,{i_k}}$, which is by convention skew-symmetric with respect to an odd permutation of subset indices. For example, a combinatorial gradient maps 1-cliques to 2-cliques
\begin{equation}
    (grad\, \phi)_{ij} = \phi_j - \phi_i,
\end{equation}
and a combinatorial curl maps 2-cliques to 3-cliques
\begin{equation}
    (grad\, \psi)_{ijk} = \psi_{ij} + \psi_{jk} + \psi_{ki}.
\end{equation}
One may also introduce adjoint operators; for example the following combinatorial divergence is adjoint to the gradient
\begin{equation}
    (grad^* \psi)_i = \sum_{j \sim i} \psi_{ij}.
\end{equation}
Hence, we may easily interpret the aggregation step of a traditional GNN as a DNN composed with the combinatorial divergence.

These operators mimic algebraic structure from the familiar vector calculus. They form an \textit{exact sequence} by guaranteeing $div\, curl = curl\, grad = 0$, and may be used to construct a graph Laplacian $\delta = grad^* \circ grad$ and higher-order Hodge Laplacians with well-characterized spectra, a Lax-Milgram theory, Poincare inequalities, and a Hodge decomposition, all of which function analogously to the familiar vector calculus setting. 

In \citep{seo2019differentiable}, the authors use this vector calculus to fit Hodge-Laplacian operators to diffusion processes on scattered data, introducing a PINN-style regularizer to fit a GNN to data while penalizing deviations from an assumed diffusion;
see also Fig. \ref{fig:DPGN}. For example, for a diffusion problem, penalizing an explicit Euler update of the form 
$\bm{x}^{n+1} = \bm{x}^n + \alpha grad^* \circ grad \bm{x}^n$ promotes diffusion, where $n$ denotes a given timestep. This process is limited to describing diffusion processes only.

In \citep{trask2022enforcing}, the graph calculus is used to learn boundary value problems associated with conservation balances, such as conservation of flux or circulation. For example, a conservation law consisting of a diffusive plus nonlinear flux may be learned
\begin{center}
\begin{tabular}{c c c}
$\mathbf{F} + \kappa \nabla \phi + N(\nabla \phi) = 0$ & $\rightarrow$ & $\mathbf{w}_1 + grad\, \mathbf{u}_0 + \mathcal{N}[grad\,\mathbf{u}_0;\theta]= 0$\\
    $\nabla \cdot \mathbf{F} = f$ & & $grad^*\, \mathbf{w}_1 = \mathbf{f}_0.$
\end{tabular}
\end{center}
Here, subscripts $0$ and $1$ denote scalar values associated with $1-$ and $2-$cliques respectively, N is a nonlinearity, and $\mathcal{N}$ denotes a trainable neural network with parameters $\theta$. An additional parameterization of a Hodge star operator provides a means of learning metric information. This generalizes the GNNs discussed previously: the first equation corresponds to an update function on edges encoding fluxes, while the second corresponds to an aggregate step enforcing conservation of flux. One arrives at an operator of the form
\begin{align}
    grad^* grad \,\mathbf{u}_0 + grad^* \mathcal{N}[u_0;\theta] = \bm{f}_0,
\end{align}
so the operator is a nonlinear perturbation of a Hodge-Laplacian. This can equivalently be given in variational form, where  the solution $\bm{u}_0$ satisfies for all $v$
\begin{equation}
    a(\bm{u}_0,v) + N_v[\bm{u}_0] = (f,v),
\end{equation}
with $a$ denoting an elliptic bilinear form. This can rigorously be shown to have a unique solution, and therefore an equality constrained optimization strategy can be applied to fit such models to data while exactly enforcing physics, independent of available data.

A key feature of this framework is the ability to handle non-trivial null-spaces which arise from involution constraints in electromagnetism. This yields an attractive framework for AI/ML-enabled data-driven modeling in microelectronics and magneto-hydrodynamics which mandate structure preservation \citep{gao2020physics,bochev2003towards}. We comment, however, that this notion of compatibility is appropriate for non-linear elliptic problems. Other notions of compatibility are important for other problems; for example, geometric mechanics is important for dynamical systems while for flow problems it is necessary to consider Lie derivatives for advection operators \citep{hernandez2022thermodynamics,lee2021machine,marsden1998multisymplectic}.

In non-physics based applications, the graph exterior calculus plays an important role in topological data analysis. For example, statistical ranking may be used to indicate preferences as a flow on a graph of options \citep{jiang2011statistical}, and the Betti numbers of a graph corresponding to the null-space of combinatorial Hodge Laplacians may be used to identify connected components and other geometric features in datasets \citep{carlsson2009topology}.




\subsection{Scalable GNNs}
There are several GNN frameworks available which utilize modern distributed and heterogeneous computing architectures to train over large graph-based data structures. The usage of these frameworks depends on i) whether to employ full-graph or mini-batch training, and ii) whether they are optimized for CPU-only or hardware accelerated computing architectures. Frameworks for distributed and full graph training for CPU and GPUs were rigorously developed in \citep{hall2021ginns,md2021distgnn, jia2020improving}, whereas for mini-batch based graph training, packages were developed and optimized for CPU-based clusters \citep{tripathy2020reducing, zhang2020agl}. It is necessary to note that full-graph training requires many epochs in comparison to mini-batch training and also converges to lower accuracy \citep{zheng2021distributed, keskar2016large, lecun2012efficient, wilson2003general}.

When considering hardware acceleration of GNNs, it is imperative to consider whether GPUs will be beneficial to the training of GNNs. For GPU acceleration of GNN training to be effective, we require a high on-GPU compute intensity relative to transfer of data between the CPU and GPU. In general, GNNs require irregular memory access and computation due to their unstructured nature. This is in contrast to training of dense networks which consist of matrix vector product kernels over layers of regular size. Therefore, for the training of large graphs, we require efficient strategies to move data from CPU to GPU and mitigate latency. 

A second challenge for implementing GNNs on distrbuted frameworks is how to effectively load balance across mini-batches. Ideally, we aim to simultaneously balance both the vertices/edges across processors and the amount of data to be transferred from memory to register. Load balancing is also required to synchronize the parameter update across all processes during backprop. 

Natural graph networks \citep{de2020natural} treat complex and heterogeneous graphs with varying nodal feature sizes. To efficiently train heterogeneous graph architectures on hybrid CPU/GPUs based clusters,  \cite{zheng2021distributed} proposed the DistDGLv2 framework, which is based on the Deep Graph Library (DGL) \citep{wang2019deep}. In this framework, mini-batch training is performed on GPUs, whereas the graph structure along with the vertices and edges are stored in CPU memory. The authors have shown a  $20\times$ speed-up for the GraphSage benchmark \citep{hamilton2017inductive} and a $36\times$ speed-up on Graph Attention Network benchmarks \citep{velickovic2018pietro} using 64 GPUs on Amazon's EC2 cluster. Similar to \citep{zeng2019accurate},  \citep{bruss2019graph} propose a distributed framework for CPU-based clusters. 

In a more general setting, scaling of unstructured graph algorithms is an important problem. Within the Department of Energy, the ExaGraph Co-design Center is a component of the Exascale Computing Project, which aims to develop broadly applicable graph kernels for exascale computation (1 exaflop = $10^{18}$ flops) in applications spanning power grid, biology, chemistry, wind energy, and national security \citep{acer2021exagraph}. The project aims to co-design algorithms, computing architecture, and hardware to build combinatorial kernels impacting general graph computation in addition to graph neural networks. 


\section{Summary and Outlook}
\label{sec:Summary}
Unlike traditional machine learning, in scientific machine learning there are not big data available, and moreover, the few measurements available may be for auxiliary variables and not
on the domain boundaries to provide proper boundary conditions. Physics-informed learning has emerged as a powerful approach in tackling such ill-defined problems and can use synergistically any available data from measurements together with data that are computed based on the physics of the problem and the corresponding governing equations. This can be done on random points as in physics-informed neural networks (PINNs) or on the nodes of a graph as in graph neural networks (GNNs).

PINNs and their extensions are relatively new methods that can tackle such problems, and they are especially effective for inverse problems, which may be challenging for traditional numerical methods. However, for realistic applications, new PINN implementations are needed that are amenable to multi-GPU computing. At the present time, a hybrid multi-GPU/multi-CPU paradigm based on domain decomposition seems to emerge as the most efficient solution but more work is required both on algorithms as well as on implementation, including mixed precision calculations, that can significantly speed up PINNs.

GNNs offer effective alternative solutions, especially for unstructured data and very complex systems. Here, we have reviewed spectral methods, spatial methods and graph attention networks that can all be used in the context of physics-informed learning, including discovering smart finite difference stencils and operators. We also reviewed physics-informed graph networks (PIGNs), where graph exterior calculus is used explicitly to construct the various differential operators on graphs, and preliminary results for simulation physical systems are very promising. In contrast to PINNs, PIGNs may impose inductive biases corresponding to exact enforcement of integral balances rather than differential operators. There are already several parallel frameworks for scaling up GNNs but the new effort should be directed towards scaling physics-informed GNNs.     

Finally, we want to comment on the connection between graphs and PDEs. It seems that the scientific machine learning community is importing ideas from graph theory and algorithms to solve PDEs, which is largely true and could be exploited even further. For example, graph embedding methods offer drastic dimensionality reduction for complex data both for static and dynamic graphs, see \citep{xu2021understanding}, hence, graph embedding could also be exploited in the context of complex dynamical systems by the PDE community.
However, recently there has been an effort to design new GNNs based on PDE theory, see, for example, the so-called ``GRAND" scheme of \citep{chamberlain2021grand}, where standard PDE tools are used to understand existing GNN architectures and subsequently develop a broader class of 
GNNs that are stable even for very deep networks. Specifically, the authors of \citep{chamberlain2021grand} have shown that many popular GNN architectures can be derived from a universal mathematical framework by choosing different forms of diffusion equations and time-discretization schemes. Hence, what we see emerging is a reciprocal ``pull-push" relationship between graph theory and algorithms on one side and PDE theory and algorithms on the other side that will be beneficial in greatly accelerating the emerging field of scientific machine learning.

\begin{Backmatter}

\paragraph{Funding}
This research was supported by the DOE project PhILMs
(no. DE- SC0019453) and the OSD/AFOSR MURI grant FA9550-20-1-0358. Trask's research was supported by the Department of Energy early career research program. Sandia National Laboratories is a multi-mission laboratory managed and operated by National Technology and Engineering Solutions of Sandia, LLC., a wholly owned subsidiary of Honeywell International, Inc., for the U.S. Department of Energy’s National Nuclear Security Administration under contract DE-NA0003525. This paper describes objective technical results and analysis.  Any subjective views or opinions that might be expressed in the paper do not necessarily represent the views of the U.S. Department of Energy or the United States Government.

\paragraph{Competing Interests}
None.

\paragraph{Data Availability Statement}
The data that support the findings of this study are openly available at https://www.sciencedirect.com/science/article/pii/S0021999121005787, https://arxiv.org/pdf/2111.02801.pdf, https://github.com/rgp62/gmls-nets and https://github.com/atzberg/gmls-nets.

\paragraph{Author Contributions}
All authors wrote and edited the paper.
All authors approved the final submitted draft.


\end{Backmatter}

\end{document}